\documentclass[iicol]{sn-jnl}% Default with double column layout
\jyear{2021}%

\theoremstyle{thmstyleone}%
\theoremstyle{thmstyletwo}%

\theoremstyle{thmstylethree}%

\usepackage{graphicx}
\usepackage{amsmath}
\usepackage{amssymb}
\usepackage{booktabs}
\usepackage{comment}
\usepackage{multirow}
\usepackage{mathtools}
\usepackage{epsfig}
\usepackage{booktabs}
\usepackage[misc]{ifsym}
\usepackage{color}

\begin{document}

\title[Article Title]{PhysFormer++: Facial Video-based Physiological Measurement with SlowFast Temporal Difference Transformer}


\author[1]{\fnm{Zitong} \sur{Yu}}%\email{zitong.yu@ieee.org}

\author[2]{\fnm{Yuming} \sur{Shen}}%\email{ym\_zmxncbv@hotmail.com}

\author[3]{\fnm{Jingang} \sur{Shi}}%\email{jingang@xjtu.edu.cn}

\author[4]{\fnm{Hengshuang} \sur{Zhao}}%\email{hszhao@cs.hku.hk}

\author[5]{\fnm{Yawen} \sur{Cui}}%\email{yawen.cui@oulu.fi}

\author[5]{\fnm{\\Jiehua} \sur{Zhang}}%\email{jiehua.zhang@oulu.fi}

\author[2]{\fnm{Philip} \sur{Torr}}%\email{philip.torr@eng.ox.ac.uk}

\author[5]{ \fnm{Guoying} \sur{Zhao \Letter}}%\email{guoying.zhao@oulu.fi}

\affil[1]{\orgname{Great Bay University}, \orgaddress{\city{Dongguan}, \postcode{523000}, \country{China}}}

\affil[2]{\orgname{University of Oxford}, \orgaddress{\city{Oxford}, \postcode{OX13PJ}, \country{UK}}}

\affil[3]{\orgname{Xi'an Jiaotong University}, \orgaddress{ \city{Xi'an}, \postcode{710049}, \country{China}}}

\affil[4]{\orgname{The University of Hong Kong}, \orgaddress{\city{Hong Kong}, \country{China}}}

\affil[5]{\orgname{University of Oulu}, \orgaddress{\city{Oulu}, \postcode{90014}, \country{Finland}}}


\abstract{Remote photoplethysmography (rPPG), which aims at measuring heart activities and physiological signals from facial video without any contact, has great potential in many applications (e.g., remote healthcare and affective computing). Recent deep learning approaches focus on mining subtle rPPG clues using convolutional neural networks with limited spatio-temporal receptive fields, which neglect the long-range spatio-temporal perception and interaction for rPPG modeling. In this paper, we propose two end-to-end video transformer based architectures, namely PhysFormer and PhysFormer++, to adaptively aggregate both local and global spatio-temporal features for rPPG representation enhancement. As key modules in PhysFormer, the temporal difference transformers first enhance the quasi-periodic rPPG features with temporal difference guided global attention, and then refine the local spatio-temporal representation against interference. To better exploit the temporal contextual and periodic rPPG clues, we also extend the PhysFormer to the two-pathway SlowFast based PhysFormer++ with temporal difference periodic and cross-attention transformers. Furthermore, we propose the label distribution learning and a curriculum learning inspired dynamic constraint in frequency domain, which provide elaborate supervisions for PhysFormer and PhysFormer++ and alleviate overfitting. Comprehensive experiments are performed on four benchmark datasets to show our superior performance on both intra- and cross-dataset testings. Unlike most transformer networks needed pretraining from large-scale datasets, the proposed PhysFormer family can be easily trained from scratch on rPPG datasets, which makes it promising as a novel transformer baseline for the rPPG community.}




\maketitle

\section{Introduction}\label{sec1}

\begin{figure}
%\vspace{-1.0em}
\centering
\includegraphics[scale=0.41]{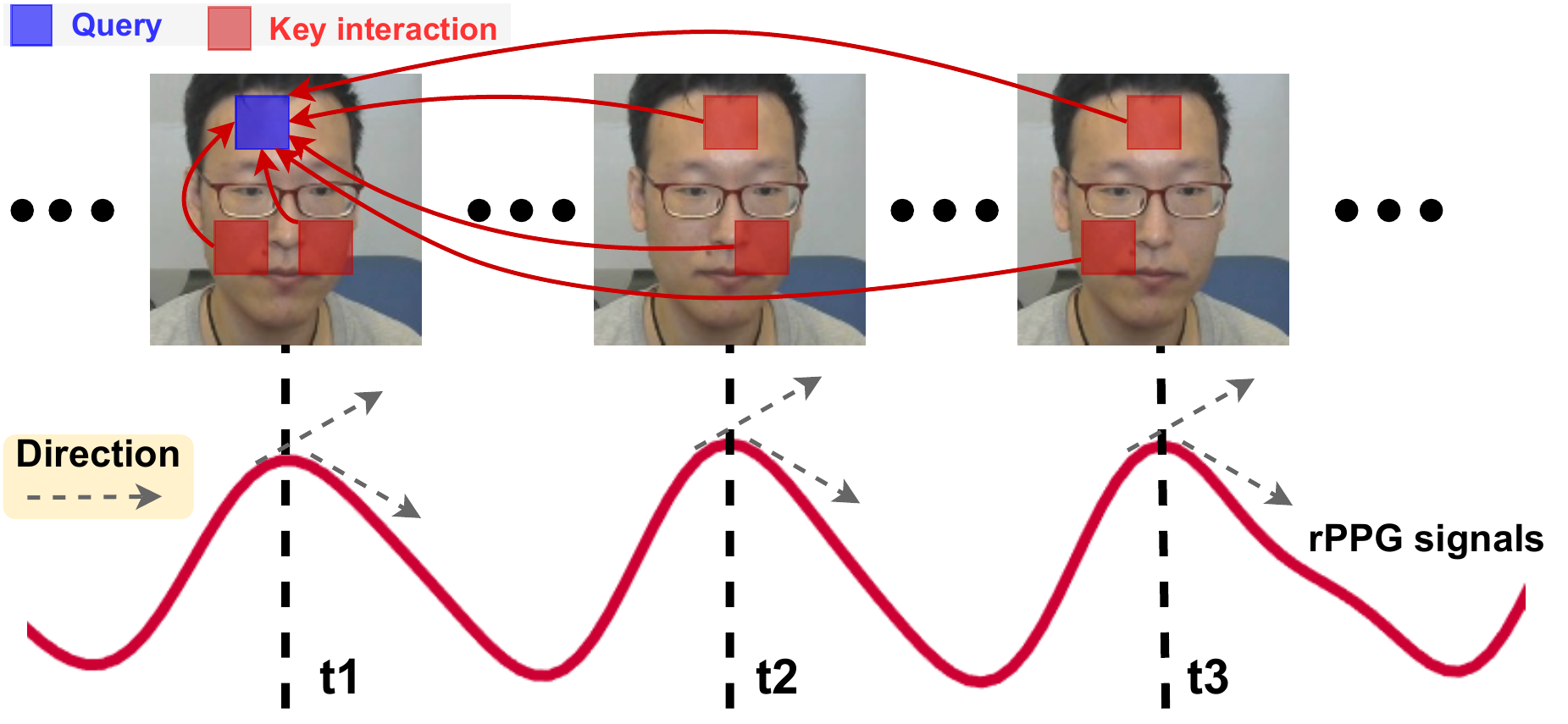}
%\includegraphics[scale=0.34]{KL.pdf}
%\vspace{-1.7em}
  \caption{\small{
  The trajectories of rPPG signals around t1, t2, and t3 share similar properties (e.g., trends with rising edge first then falling edge later, and relatively high magnitudes) induced by skin color changes. It inspires the long-range spatio-temporal attention (e.g., \textcolor{blue}{blue} tube around t1 interacted with \textcolor{red}{red} tubes from intra- and inter-frames) according to their local temporal difference features for quasi-periodic rPPG enhancement. Here `tube' indicates the same regions across short-time consecutive frames.}
  }
 
\label{fig:Figure1}
%\vspace{-1.2em}
\end{figure}

Physiological signals such as heart rate (HR), respiration frequency (RF), and heart rate variability (HRV) are important vital signs to be measured in many circumstances, especially for healthcare or medical purposes. Traditionally, the electrocardiography (ECG) and photoplethysmograph (PPG) or blood volume pulse (BVP) are the two most common ways for measuring heart activities and corresponding physiological signals. However, both ECG and PPG/BVP sensors need to be attached to body parts, which may cause discomfort and are inconvenient for long-term monitoring. To counter this issue, remote photoplethysmography (rPPG)~\cite{yu2021facial,chen2018video,liu2021camera} methods are developing fast in recent years, which aim to measure heart activity remotely without any contact.

In earlier studies of facial rPPG measurement, most methods analyze subtle color changes on facial regions of interest (ROI) with classical signal processing approaches~\cite{verkruysse2008remote,poh2010non,poh2010advancements,li2014remote,tulyakov2016self,magdalena2018sparseppg}. Besides, there are a few color subspace transformation methods~\cite{de2013robust,wang2017algorithmic} which utilize all skin pixels for rPPG measurement. Based on the prior knowledge from traditional methods, a few learning based approaches~\cite{ hsu2017deep,qiu2018evm,niu2018synrhythm,niu2019rhythmnet} are designed as non-end-to-end fashions. ROI based preprocessed signal representations (e.g., time-frequency map~\cite{ hsu2017deep} and spatio-temporal map~\cite{niu2018synrhythm,niu2019rhythmnet}) are generated first, and then learnable models could capture rPPG features from these maps. However, these methods need the strict preprocessing procedure and neglect the global contextual clues outside the pre-defined ROIs. Meanwhile, more and more end-to-end deep learning based rPPG methods~\cite{vspetlik2018visual,chen2018deepphys,yu2019remote1,yu2019remote2,liu2020multi} are developed, which treat facial video frames as input and predict rPPG and other physiological signals directly. However, pure end-to-end methods are easily influenced by the complex scenarios (e.g., with head movement and various illumination conditions) and rPPG-unrelated features can not be ruled out in learning, resulting in huge performance drops~\cite{yu2020autohr} in realistic datasets (e.g., VIPL-HR~\cite{niu2019rhythmnet}). 

Recently, due to its excellent long-range attentional modeling capacities in solving sequence-to-sequence issues, transformer~\cite{lin2021survey,han2020survey} has been successfully applied in many artificial intelligence tasks such as natural language processing (NLP)~\cite{vaswani2017attention}, image~\cite{dosovitskiy2020image} and video~\cite{bertasius2021space} analysis. Similarly, rPPG measurement from facial videos can be treated as a video sequence to signal sequence problem, where the long-range contextual clues should be exploited for semantic modeling. As shown in Fig.~\ref{fig:Figure1}, rPPG clues from different skin regions and temporal locations (e.g., signal trajectories around t1, t2, and t3) share similar properties (e.g., trends with rising edge first then falling edge later and relative high magnitudes), which can be utilized for long-range feature modeling and enhancement. However, different from the most video tasks aiming at semantic motion representation, facial rPPG measurement focuses on capturing subtle skin color changes, which makes it challenging for global spatio-temporal perception. Besides, the rPPG measurement task usually relies on periodic hidden visual dynamics, and the existing deep end-to-end models are weak in representing such clues. Furthermore, video-based rPPG measurement is usually a long-time monitoring task, and it is challenging to design and train transformers with long video sequence inputs.

Motivated by the discussions above, we propose two end-to-end video transformer architectures, namely PhysFormer and PhysFormer++, for remote physiological measurement. On the one hand, the cascaded temporal difference transformer blocks in PhysFormer benefit the rPPG feature enhancement via global spatio-temporal attention based on the fine-grained temporal skin color differences. Furthermore, the two-pathway SlowFast temporal difference transformer based PhysFormer++ with periodic- and cross-attention is able to efficiently capture the temporal contextual and periodic rPPG clues from facial videos. On the other hand, to alleviate the interference-induced overfitting issue and complement the weak temporal supervision signals, elaborate supervision in frequency domain is designed, which helps the PhysFormer family learn more intrinsic rPPG-aware features. 
%Our contributions include:

This paper is an extended version of our prior work~\cite{yu2021physformer} accepted by CVPR 2022. The main differences with the conference version are as follows: 1) besides the temporal difference transformer based PhysFormer, we propose the novel SlowFast video transformer architecture PhysFormer++ for rPPG measurement task; 2) based on the temporal difference transformer, the temporal difference periodic transformer and temporal difference cross-attention transformer are proposed to enhance the rPPG periodic perception and cross-tempo rPPG dynamics, respectively; 3) a detailed overview about the traditional, non-end-to-end learning based, and end-to-end learning based rPPG measurement methods is discussed in the related work; 4) more elaborate experimental results, visualization, and efficiency analysis are given for the PhysFormer family. To sum up, the main contributions of this paper are listed:
%\vspace{-0.5em}
\begin{itemize}
\setlength\itemsep{-0.1em}
%\vspace{-0.5em}
    
    \item We propose the PhysFormer family, i.e., PhysFormer and PhysFormer++, which mainly consists of a powerful video temporal difference transformer backbone. To our best knowledge, it is the first time to explore the long-range spatio-temporal relationship for reliable rPPG measurement. Besides, the proposed temporal difference transformer is potential for broader fine-grained or periodic video understanding tasks in computer vision (e.g., video action recognition and repetition counting) due to its excellent spatio-temporal representation capacity with local temporal difference description and global spatio-temporal modeling.
    
    \item We propose the two-pathway SlowFast architecture for PhysFormer++ to efficiently leverage both fine-grained and semantic tempo rPPG clues. Specifically, the temporal difference periodic and cross-attention transformers are respectively designed for the Slow and Fast pathways to enhance the representation capacity of the periodic rPPG dynamics. 

    \item We propose an elaborate recipe to supervise PhysFormer with label distribution learning and curriculum learning guided dynamic loss in frequency domain to learn efficiently and alleviate overfitting. Such curriculum learning guided dynamic strategy could benefit not only the rPPG measurement task but also general deep learning tasks such as multi-task learning and multi-loss adjusting.

    \item We conduct intra- and cross-dataset testings and show that the proposed PhysFormer achieves superior or on par state-of-the-art performance without pretraining on large-scale datasets like ImageNet-21K.
    
\end{itemize}

In the rest of the paper, Section~\ref{sec:relatedwork} provides the related work about rPPG measurement and vision transformer. Section~\ref{sec:method} first introduces the detailed architectures of the PhysFormer and PhysFormer++, and then formulates the label distribution learning and curriculum learning guided dynamic supervision for rPPG measurement. Section~\ref{sec:experiment} introduces the four rPPG benchmark datasets and evaluation metrics, and provides rigorous ablation studies, visualizations and evaluates the performance of the proposed models. Finally, a conclusion is given in Section~\ref{sec:conclusion}.

%%\vspace{-1.5em}
\section{Related Work}
\label{sec:relatedwork}

In this section, we provide a brief discussion of the related facial rPPG measurement approaches. As shown in Table~\ref{tab:rPPGmethods}, these approaches can be generally categorized into traditional, non-end-to-end learning, and end-to-end learning based methods. We also briefly review the transformer architectures for vision tasks.

\newcommand{\tabincell}[2]{\begin{tabular}{@{}#1@{}}#2\end{tabular}}
\begin{table*}
\centering
\caption{Summary of the representative rPPG measurement methods in terms of traditional, non-end-to-end learning, and end-to-end learning categories.} \label{tab:rPPGmethods}
\resizebox{0.99\textwidth}{!} {\begin{tabular}{l| c c c c c} 
% \toprule[1pt]
\midrule
 & Method & Venue & Feature & Backbone & Loss Function \\
 
\midrule
\multirow{14}{*}{\tabincell{c}{\quad\rotatebox{90}{\textbf{Traditional}}}\quad}& \tabincell{c}{ Poh2010}~\cite{poh2010advancements} & \tabincell{c}{IEEE Trans. \\ Biomed. Eng.} & \tabincell{c}{ROI selection, ICA decomposition} & \tabincell{c}{-} & \tabincell{c}{-}  \\ 

\cmidrule{2-6}
& \tabincell{c}{CHROM}~\cite{de2013robust} & \tabincell{c}{IEEE Trans. \\ Biomed. Eng.} & \tabincell{c}{Mapping on the color
difference (chrominance) subspace} & - & -   \\

\cmidrule{2-6}

& \tabincell{c}{Li2014}~\cite{li2014remote} & CVPR & \tabincell{c}{ROI selection, tracking, illumination rectification, \\ non-rigid motion elimination} & - & -   \\ 

\cmidrule{2-6}
& \tabincell{c}{RandomPatch}~\cite{lam2015robust} & ICCV & \tabincell{c}{ICA decomposition on random patches, majority voting} & - & -   \\ 

\cmidrule{2-6}
& \tabincell{c}{Tulyakov2016}~\cite{tulyakov2016self} & CVPR & \tabincell{c}{ Chrominance features from multiple ROIs,\\low-rank factorization via self-adaptive matrix completion } & - & -   \\ 

\cmidrule{2-6}
& \tabincell{c}{POS}~\cite{wang2017algorithmic} & \tabincell{c}{IEEE Trans. \\ Biomed. Eng.} & \tabincell{c}{Mapping on the projection
plane orthogonal to the skin tone} & - & -   \\

\midrule
\multirow{14}{*}{\tabincell{c}{\quad\rotatebox{90}{\textbf{Non-end-to-end Learning}}}\quad}& \tabincell{c}{SynRhythm}~\cite{niu2018synrhythm} & ICPR & \tabincell{c}{spatio-temporal map representation, pretrained on Synthetic Rhythms} & \tabincell{c}{ResNet18} & \tabincell{c}{L1 regression loss}  \\ 

\cmidrule{2-6}
& \tabincell{c}{EVM-CNN}~\cite{qiu2018evm} & IEEE TMM & \tabincell{c}{ROI tracking, feature image via spatial decomposition\\ and temporal filtering} & \tabincell{c}{Shallow CNN} & \tabincell{c}{L2 regression loss}  \\ 

\cmidrule{2-6}
& \tabincell{c}{RhythmNet}~\cite{niu2019rhythmnet} & IEEE TIP & \tabincell{c}{spatio-temporal map in YUV color space,\\ temporal contextual modeling via RNN} & \tabincell{c}{ResNet18+GRU} & \tabincell{c}{L1 regression loss\\temporal smooth loss}  \\ 

\cmidrule{2-6}
& \tabincell{c}{ ST-Attention}~\cite{niu2019robust} & IEEE FG & \tabincell{c}{spatio-temporal map in YUV color space, channel and\\ spatio-temporal attention, temporal augmentation} & \tabincell{c}{ResNet18} & \tabincell{c}{L1 regression loss}  \\

\cmidrule{2-6}
& \tabincell{c}{NAS-HR}~\cite{lu2021hr} & VRIH & \tabincell{c}{Spatial–temporal map with POS signals,\\ NAS for efficient architecture search} & \tabincell{c}{lightweight NAS} & \tabincell{c}{L1 regression loss}  \\

\cmidrule{2-6}
& \tabincell{c}{CVD}~\cite{niu2020video} & ECCV & \tabincell{c}{Multi-scale spatio-temporal map on both RGB and YUV color\\ spaces, cross-verified feature disentangling, multi-task learning} & \tabincell{c}{ResNet18 with\\
shallow decoder} & \tabincell{c}{L1 regression loss, CE loss, \\NegPearson loss, reconstruction loss}  \\

\cmidrule{2-6}
& \tabincell{c}{Dual-GAN}~\cite{lu2021dual} & CVPR & \tabincell{c}{spatio-temporal map, dual GAN for\\ BVP signal and noise modeling, respectively} & \tabincell{c}{Customized CNN} & \tabincell{c}{L1 regression loss, CE loss, \\NegPearson loss, GAN loss}  \\

\midrule
\multirow{20}{*}{\tabincell{c}{\quad\rotatebox{90}{\textbf{End-to-end Learning}}}\quad}& \tabincell{c}{DeepPhys}~\cite{chen2018deepphys} & \tabincell{c}{ECCV} & \tabincell{c}{Motion representation with normalized frame difference,\\attention
mechanism using appearance to guide motion} & \tabincell{c}{Two-branch CNN} & \tabincell{c}{MSE loss}  \\

\cmidrule{2-6}
& \tabincell{c}{HR-CNN}~\cite{vspetlik2018visual} & \tabincell{c}{BMVC} & \tabincell{c}{Two-stage CNN to measure the rPPG signals first,\\ and then estimate the HR value} & \tabincell{c}{Shallow CNN} & \tabincell{c}{SNR loss, L1 regression loss}  \\ 

\cmidrule{2-6}
& \tabincell{c}{PhysNet}~\cite{yu2019remote1} & BMVC & \tabincell{c}{End-to-end spatio-temporal networks} & \tabincell{c}{3DCNN} & \tabincell{c}{NegPearson loss}  \\ 

\cmidrule{2-6}
& \tabincell{c}{rPPGNet}~\cite{yu2019remote2} & ICCV & \tabincell{c}{Two-stage framework to enhance video\\ quality first, and then extract rPPG signals} & \tabincell{c}{3DCNN} & \tabincell{c}{NegPearson loss, reconstruction loss\\ skin segmentation loss}  \\ 

\cmidrule{2-6}
& \tabincell{c}{AutoHR}~\cite{yu2020autohr} & IEEE SPL & \tabincell{c}{NAS with temporal difference convolution,\\ spatio-temporal augmentation} & \tabincell{c}{NAS 3DCNN} & \tabincell{c}{NegPearson loss, CE loss}  \\

\cmidrule{2-6}
& \tabincell{c}{TS-CAN}~\cite{liu2020multi} & NeuIPS & \tabincell{c}{Multi-task temporal shift convolutional attention networks,\\ mobile-level real-time rPPG and respiratory measurement} & \tabincell{c}{Two-branch \\temporal shift CNN} & \tabincell{c}{MSE loss on\\ pulse and respiration}  \\

\cmidrule{2-6}
& \tabincell{c}{EfficientPhys}~\cite{liu2021efficientphys} & arXiv & \tabincell{c}{Temporal difference normalization, \\self-attention-shifted networks} & \tabincell{c}{Temporal shift\\ Swin Transformer} & \tabincell{c}{MSE loss}  \\ 

\cmidrule{2-6}
& \tabincell{c}{PhysFormer (Ours)\\PhysFormer++ (Ours)} & 2022 & \tabincell{c}{Temporal difference transformers supervised by the\\ label distribution learning and curriculum learning strategy} & \tabincell{c}{Temporal difference \\video transformer} & \tabincell{c}{NegPearson loss, CE loss,\\label distribution loss}  \\

 %\hline
 \bottomrule[1pt]
 \end{tabular}}
\end{table*}

\subsection{rPPG measurement}

\noindent\textbf{Traditional approaches.}\quad      
An early study of rPPG-based physiological measurement was reported in~\cite{verkruysse2008remote}. Plenty of traditional hand-crafted approaches have been developed in this field since then. Compared with coarsely averaging arbitrary color channel from the detected full face region, selective merging information from different color channels~\cite{poh2010non,poh2010advancements} from different ROIs~\cite{lam2015robust,li2014remote} with adaptive temporal filtering~\cite{li2014remote} are proven to be more efficient for subtle rPPG signal recovery. To improve the signal-to-noise rate of the recovered rPPG signals, several signal decomposition methods such as independent component analysis (ICA)~\cite{poh2010non,poh2010advancements,lam2015robust} and matrix completion~\cite{tulyakov2016self} are also proposed. To alleviate the impacts of the skin tone and head motion, several color space projection (e.g., chrominance subspace~\cite{de2013robust} and skin-orthogonal space~\cite{wang2017algorithmic}) methods are developed. Despite remarkable early-stage progresses, these approaches have the following limitations: 1) they
require empirical knowledge to design the components (e.g., hyperparameter in signal processing filtering); 2) there is a lack of supervised learning models to counter data variations, especially in challenging environments with serious interference.

\vspace{0.5em}
\noindent\textbf{Non-end-to-end learning approaches.}\quad 
In recent years, deep learning based approaches dominate the field of rPPG measurement due to the strong spatio-temporal representation capabilities. One representative framework is to learn robust rPPG features from the facial ROI-based spatio-temporal signal map (STmap). STmap~\cite{niu2018synrhythm,niu2019robust} or its variants (e.g., multisacle STmap~\cite{niu2020video,lu2021dual} and chrominance STmap~\cite{lu2021hr}) are first extracted from predefined facial ROIs on different color spaces, and then classical convolutional neural network (CNN) (e.g., ResNet~\cite{he2016deep}) and recurrent neural network (RNN) (e.g., GRU~\cite{cho2014properties}) are cascaded for rPPG feature representation. The STmap-based non-end-to-end learning framework focuses on learning an underlying mapping from the input feature maps to the target rPPG signals. With dense raw rPPG information and less irrelevant elements (e.g., face-shape attributes), these methods usually converge faster and achieve reasonable performance against head movement but need explicit and exhaustive preprocessings.

\vspace{0.5em}
\noindent\textbf{End-to-end learning approaches.}\quad  
Besides learning upon handcrafted STmaps, end-to-end learning from facial sequence directly is also favorite. Both spatial 2DCNN networks ~\cite{vspetlik2018visual,chen2018deepphys} and spatio-temporal models~\cite{yu2019remote1,yu2019remote2,yu2020autohr,liu2020multi,liu2021efficientphys,nowara2021benefit,gideon2021way} are developed for rPPG feature representation. Yu et al.~\cite{yu2019remote1} investigates the recurrent methods (PhysNet-LSTM, PhysNet-ConvLSTM) for rPPG measuremnt. However, such CNN+LSTM based architectures are good at long-range sequential modeling via LSTM but fail to explore long-range intra-frame spatial relationship using CNN with local convolutions. In contrast, with spatial transformer backbone and temporal shift module, EfficientPhys~\cite{liu2021efficientphys} is able to explore long-range spatial but only short-term temporal relationship. In other words, existing end-to-end methods only consider the spatio-temporal rPPG features from local neighbors and adjacent frames but neglect the long-range relationship among quasi-periodic rPPG features. 

Compared with the non-end-to-end learning based methods, end-to-end approaches are less dependent on task-related prior knowledge and handcrafted engineering (e.g., STmap generation) but rely on diverse and large-scale data to alleviate the problem of overfitting. To enhance the long-range contextual spatio-temporal representation capacities and alleviate the data-hungry requirement of the deep rPPG models, we propose the PhysFormer and PhysFormer++ architectures, which can be easily trained from scratch on rPPG datasets with the elaborate supervision recipe.

\begin{figure*}
\centering
\includegraphics[scale=0.44]{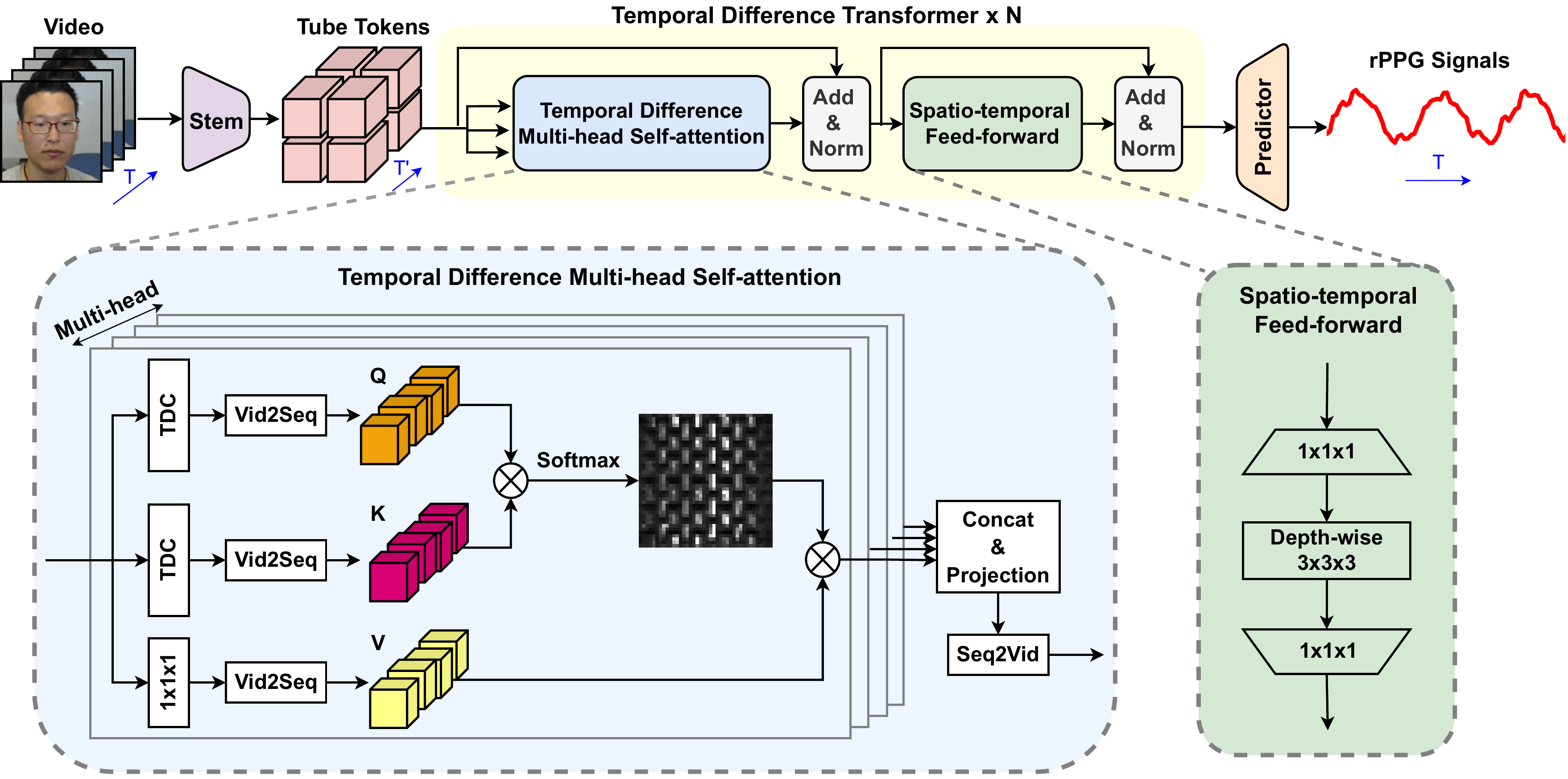}
%\vspace{-0.8em}
  \caption{\small{
  Framework of the PhysFormer. It consists of a shallow stem, a tube tokenizer, several temporal difference transformers, and a rPPG predictor head. The temporal difference transformer is formed from the Temporal Difference Multi-head Self-attention (TD-MHSA) and Spatio-temporal Feed-forward (ST-FF) modules, which enhances the global and local spatio-temporal representation, respectively. `TDC' is short for the temporal difference convolution~\cite{yu2020autohr,yu2021searching}.
  }
  }
\label{fig:PhysFormer}
%\vspace{-1.0em}
\end{figure*}

\subsection{Transformer for vision tasks}
Due to the powerful self-attention based long-range modeling capacity, transformer~\cite{lin2021survey,vaswani2017attention} has been successfully applied in the field of NLP to model the contextual relationship for sequential data. Then vision transformer (ViT)~\cite{dosovitskiy2020image} is proposed recently by feeding transformer with sequences of image patches for image classification. Many other ViT variants ~\cite{han2020survey,khan2021transformers,touvron2021training,liu2021swin,yuan2021tokens,wang2021pyramid,han2021transformer,chen2021crossvit,ding2021hr} are proposed from then, which achieve promising performance compared with its counterpart CNNs for image analysis tasks~\cite{carion2020end,zheng2021rethinking,he2021transreid}. 
Recently, some works introduce vision transformer for video understanding tasks such as action recognition~\cite{arnab2021vivit,fan2021multiscale,neimark2021video,girdhar2019video,liu2021video,bulat2021space,bertasius2021space}, action detection~\cite{zhao2021tuber,liu2021end,wang2021temporal,xu2021long}, video super-resolution~\cite{cao2021video}, video inpainting~\cite{zeng2020learning,liu2021fuseformer}, and 3D animation~\cite{chen2021geometry,chen2021aniformer}. Some works~\cite{neimark2021video,girdhar2019video} conduct temporal contextual modeling with transformer based on single-frame features from pretrained 2D networks, while other works~\cite{bertasius2021space,arnab2021vivit,liu2021video,bulat2021space,fan2021multiscale} mine the spatio-temporal attentions via video transformer directly. Most of these works are incompatible for long-video-sequence ($\textgreater$150 frames) signal regression task. There are two related works~\cite{yu2021transrppg,liu2021efficientphys} using ViT for rPPG feature representation. TransRPPG~\cite{yu2021transrppg} extracts rPPG features from the preprocessed signal maps via ViT for face 3D mask presentation attack detection~\cite{yu2021deep}. Based on the temporal shift networks~\cite{liu2020multi,lin2019tsm}, EfficientPhys-T~\cite{liu2021efficientphys} adds several Swin Transformer~\cite{liu2021swin} layers for global spatial attention. Different from these two works, the proposed PhysFormer and PhysFormer++ are end-to-end video transformers, which are able to capture long-range spatio-temporal attentional rPPG features from facial video directly.

\section{Methodology}
\label{sec:method}

We will first introduce the architecture of PhysFormer and PhysFormer++ in Sec.~\ref{sec:PhysFormer} and ~\ref{sec:PhysFormer++}, respectively. Then we will introduce label distribution learning for rPPG measurement in Sec.~\ref{sec:distribution}, and at last present the curriculum learning guided dynamic supervision in Sec.~\ref{sec:dynamic}.

\subsection{PhysFormer}
\label{sec:PhysFormer}

As illustrated in Fig.~\ref{fig:PhysFormer}, PhysFormer consists of a shallow stem $\mathbf{E}_{\text{stem}}$, a tube tokenizer $\mathbf{E}_{\text{tube}}$, $N$ temporal difference transformer blocks $\mathbf{E}^{i}_{\text{trans}}$ ($ i=1,...,N$) and a rPPG predictor head. Inspired by the study in~\cite{xiao2021early}, we adopt a shallow stem
to extract coarse local spatio-temporal features, which benefits the fast convergence and clearer subsequent global self-attention. Specifically, the stem is formed by three convolutional blocks with kernel size (1x5x5), (3x3x3) and (3x3x3), respectively. Each convolution operator is cascaded with a batch normalization (BN), ReLU and MaxPool. The pooling layer only halves the spatial dimension. Therefore, given an RGB facial video input $X\in \mathbb{R}^{3\times T\times H\times W}$, the stem output $X_{\text{stem}}=\mathbf{E}_{\text{stem}}(X)$, where $X_{\text{stem}}\in \mathbb{R}^{D\times T\times H/8\times W/8}$, and $D$, $T$, $W$, $H$ indicate channel, sequence length, width, height, respectively. Then $X_{\text{stem}}$ would be partitioned into spatio-temporal tube tokens $X_{\text{tube}}\in \mathbb{R}^{D\times T'\times H'\times W'}$ via the tube tokenizer $\mathbf{E}_{\text{tube}}$. Subsequently, the tube tokens will be forwarded with $N$ temporal difference transformer blocks and obtain the global-local refined rPPG features $X_{\text{trans}}$, which has the same dimensions with $X_{\text{tube}}$. Finally, the rPPG predictor head temporally upsamples, spatially averages, and projects the features $X_{\text{trans}}$ to 1D signal $Y\in \mathbb{R}^{T}$.

\vspace{0.4em}
\noindent\textbf{Tube tokenization.}\quad  
 Here the coarse feature $X_{\text{stem}}$ would be partitioned into non-overlapping tube tokens via $\mathbf{E}_{\text{tube}}(X_{\text{stem}})$, which aggregates the spatio-temporal neighbor semantics within the tube region and reduces computational costs for the subsequent transformers. Specifically, the token tokenizer consists of a learnable 3D convolution with the same kernel size and stride (non-overlapping setting) as the targeted tube size $T_{s}\times H_{s}\times W_{s}$. Thus, the expected tube token map $X_{\text{tube}}\in \mathbb{R}^{D\times T'\times H'\times W'}$ has length, height and width
\begin{equation} 
T'=\left \lfloor \frac{T}{T_{s}} \right \rfloor, H'=\left \lfloor \frac{H/8}{H_{s}} \right \rfloor, W'=\left \lfloor \frac{W/8}{W_{s}} \right \rfloor.
\label{eq:token}
%%\vspace{-2.1em}
\end{equation}
Please note that there are no positional embeddings after the tube tokenization as the stem with cascaded convolutions and poolings at early stage already captures relative spatio-temporal positional information~\cite{hassani2021escaping}.

\begin{figure*}
%\vspace{-1.0em}
\centering
\includegraphics[scale=0.6]{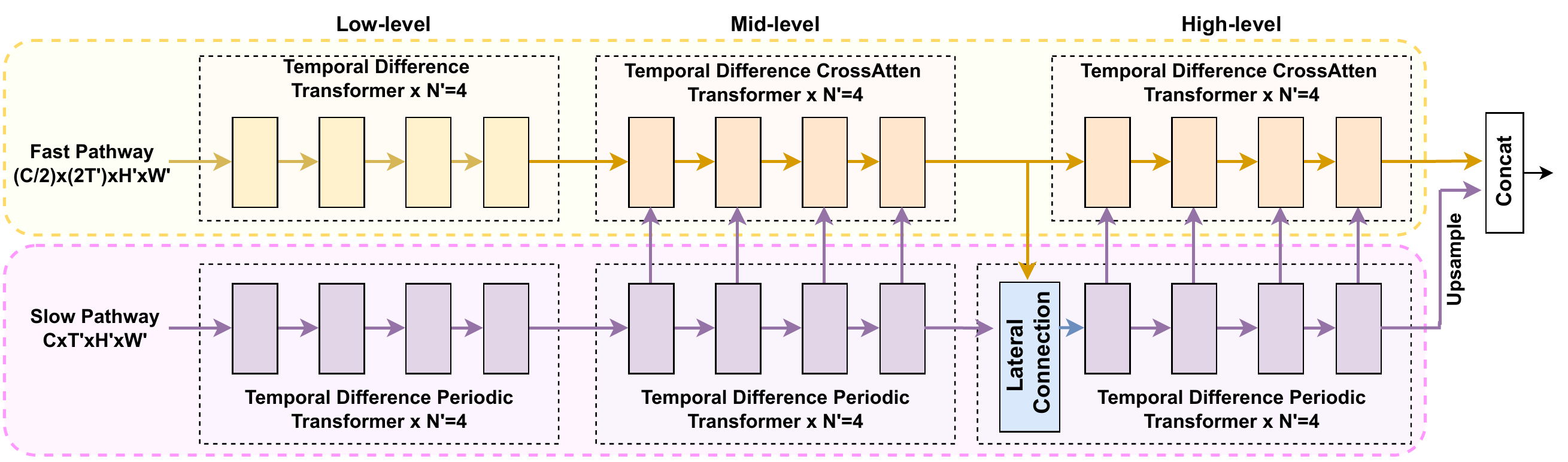}
%\includegraphics[scale=0.34]{KL.pdf}
%\vspace{-1.7em}
  \caption{\small{
  Framework of the PhysFormer++ with two-stream SlowFast pathways. Different from the PhysFormer using only slow pathway, the PhysFormer++ extracts and fuses attentional features from slow and fast pathways. Moreover, temporal difference periodic transformer blocks are used in the slow pathway. The information flow between two pathways interacts via temporal difference cross-attention transformer blocks and lateral connection.}
  }
 
\label{fig:SlowFast}
%\vspace{-1.2em}
\end{figure*}

\vspace{0.4em}
\noindent\textbf{Temporal difference multi-head self-attention TD-MHSA)}.\quad  
In self-attention mechanism~\cite{vaswani2017attention,dosovitskiy2020image}, the relationship between the tokens is modeled by the similarity between the projected query-key pairs, yielding the attention score. Instead of point-wise linear projection, we utilize temporal difference convolution (TDC)~\cite{yu2020autohr,yu2021searching} for query ($Q$) and key ($K$) projection, which could capture fine-grained local temporal difference features for subtle color change description. 
TDC with learnable $w$ can be formulated as
\begin{equation} \footnotesize
\begin{split}
\mathrm{TDC}(x)
&=\underbrace{\sum_{p_n\in \mathcal{R}}w(p_n)\cdot x(p_0+p_n)}_{\text{vanilla 3D convolution}}+\theta\cdot (\underbrace{-x(p_0)\cdot\sum_{p_n\in \mathcal{R'}}w(p_n))}_{\text{temporal difference term}}, \\
\label{eq:CDC-T}
\end{split}
\end{equation}
where $p_0$=(0,0,0) indicates the current spatio-temporal location. $\mathcal{R}$= $\left \{  (-1,-1,-1),(-1,-1,0),\cdots,(0,1,1),(1,1,1) \right \}$ indicates the sampled local (3x3x3) spatio-temporal receptive field cube for 3D convolution in both current ($t_{0}$) and adjacent time steps ($t_{-1}$ and $t_{1}$), while $\mathcal{R'}$ only indicates the local spatial regions in the adjacent time steps (t-1 and t1). The hyperparameter $\theta \in$[0, 1] tradeoffs the contribution of temporal difference. The higher value of $\theta$ means the more importance of temporal difference information (e.g., trends of the skin color changes). Specially, TDC degrades to vanilla 3D convolution when $\theta$= 0. Then query and key are projected via unshared TDC and BN as
%\vspace{-0.3em}
\begin{equation} 
Q = \mathrm{BN}(\mathrm{TDC}(X_{\text{tube}})), K= \mathrm{BN}(\mathrm{TDC}(X_{\text{tube}})). 
%\vspace{-0.3em}
\end{equation}
For the value ($V$) projection, point-wise linear projection without BN is utilized. Then $Q,K,V\in \mathbb{R}^{D\times T'\times H'\times W'}$ are flattened into sequence, and separated into $h$ heads ($D_h=D/h$ for each head). For the $i$-th head ($i\leq h$), the self-attention (SA) can be formulated
%\vspace{-0.3em}
\begin{equation}
\mathrm{SA}_{i}=\mathrm{Softmax}(Q_{i}K^{T}_{i}/\tau)V_{i},
%\vspace{-0.3em}
\end{equation}
where $\tau$ controls the sparsity. We find that the default setting $\tau=\sqrt{D_h}$ in~\cite{vaswani2017attention,dosovitskiy2020image} performs poorly for rPPG measurement. According to the periodicity of rPPG features, we use a smaller $\tau$ value to obtain sparser attention activation. The corresponding study can be found in Table~\ref{tab:ablation2}. The output of TD-MHSA is the concatenation of SA from all heads and then with a linear projection $U\in \mathbb{R}^{D\times D}$
%\vspace{-0.3em}
\begin{equation} 
\text{TD-MHSA} = \mathrm{Concat}(\mathrm{SA}_{1}; \mathrm{SA}_{2};...; \mathrm{SA}_{h})U.
%\vspace{-0.3em}
\end{equation}
As illustrated in Fig.~\ref{fig:PhysFormer}, residual connection and layer normalization (LN)  would be conducted after TD-MHSA.

\vspace{0.4em}
\noindent\textbf{Spatio-temporal feed-forward (ST-FF)}.
The vanilla feed-forward network consists of two linear transformation layers, where the hidden dimension $D'$ between two layers is expanded to learn a richer feature representation. In contrast, we introduce a depthwise 3D convolution (with BN and nonlinear activation) between these two layers with extra slight computational cost but remarkable performance improvement. The benefits are two-fold: 1) as a complementation of TD-MHSA, ST-FF could refine the local inconsistency and parts of noisy features; 2) richer locality provides TD-MHSA sufficient relative position cues.

\subsection{PhysFormer++}
\label{sec:PhysFormer++}

In the PhysFormer, the temporal length $T_{s}$ of the tube token map is fixed. However, the fixed value of $T_{s}$ might be sub-optimal for robust rPPG feature representation as the larger $T_{s}$ reduces the temporal redundancy but loses the fine-grained temporal clues, and vice versa for the smaller $T_{s}$. To alleviate this issue, we design the temporal enhanced version PhysFormer++ (see Fig.~\ref{fig:SlowFast}) consisting of two-stream SlowFast pathways with large and small $T_{s}$, respectively. Similar to the SlowFast concept in~\cite{feichtenhofer2019slowfast,kazakos2021slow}, the Slow pathway has high channel capacity with low framerates, and reduces the temporal redundancy. In contrast, the Fast pathway operates at a fine-grained temporal resolution with high framerates. Furthermore, two novel transformer blocks, \textit{temporal difference periodic transformer} and \textit{temporal difference cross-attention transformer}, are designed for the slow and fast pathway, respectively. The former one encodes contextual rPPG periodicity clues for the slow pathway while the latter one introduces efficient SlowFast interactive attentions for the fast pathway. The SlowFast architecture is able to adaptively mine richer temporally rPPG contexts for robust rPPG measurement.

\begin{figure}
%\vspace{-1.0em}
\centering
  \caption{\small{
  Architectures of PhysFormer++. Inside the brackets are the filter sizes and feature dimensionalities. `Conv' suggests the vanilla 3D convolution. All convolutional layers (except Tokenizer) are with stride=1 and are followed by a BN-ReLU layer while `MaxPool' layers are with stride=1x2x2.}
  }
\includegraphics[scale=0.43]{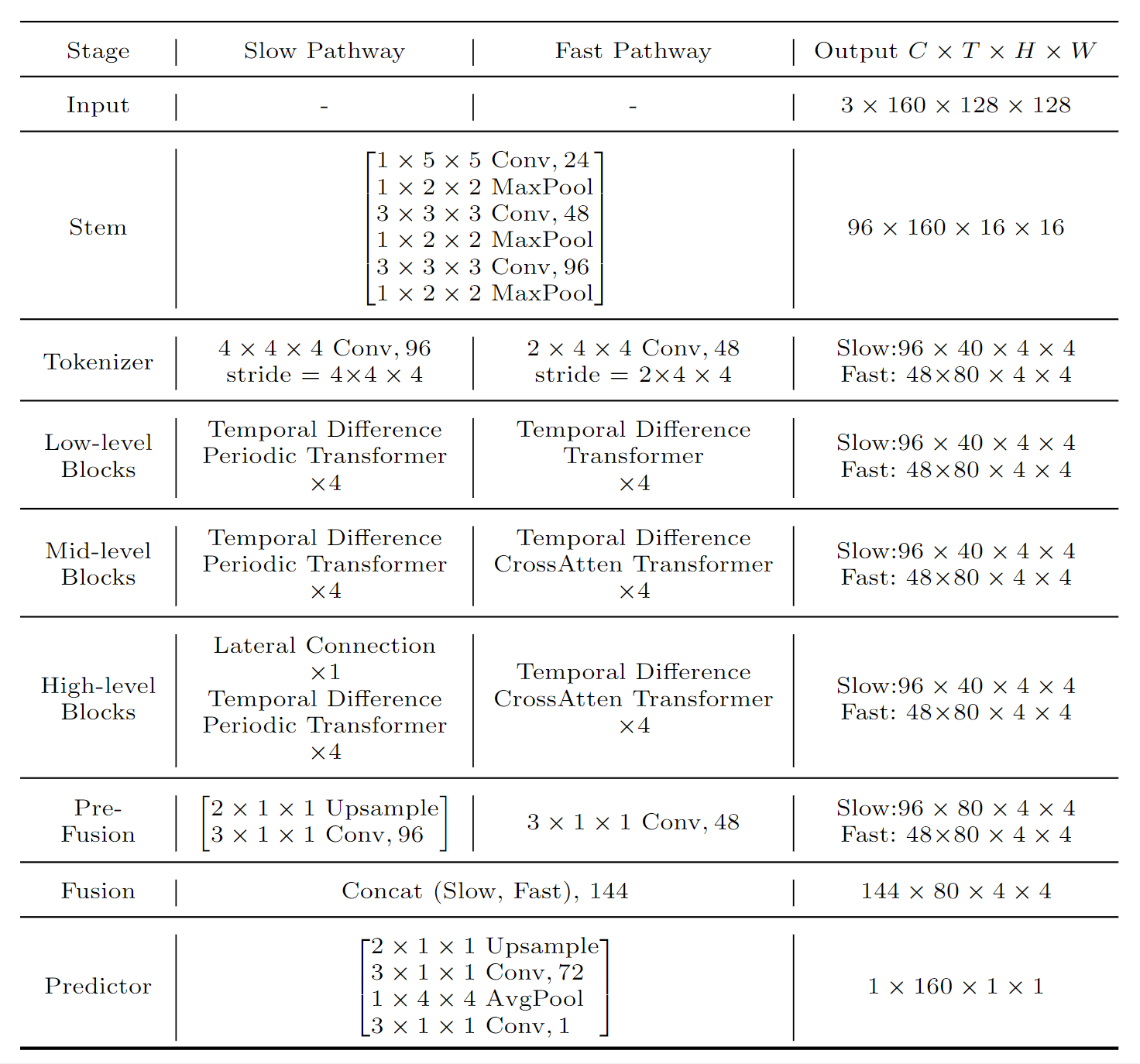}
%\includegraphics[scale=0.34]{KL.pdf}
%\vspace{-1.7em}
\label{fig:achi}
\vspace{-1.2em}
\end{figure}

As illustrated in Fig.~\ref{fig:SlowFast} and detailed architecture in Fig.~\ref{fig:achi}, different from the PhysFormer using a single tube tokenizer, two tube tokenizers $\mathbf{E}^{\text{fast}}_{\text{tube}}$ and $\mathbf{E}^{\text{slow}}_{\text{tube}}$ are adopted in PhysFormer++ to form the spatio-temporal tube tokens $X^{\text{fast}}_{\text{tube}}\in \mathbb{R}^{D^{\text{fast}}\times T^{\text{fast}}\times H'\times W'}$ and $X^{\text{slow}}_{\text{tube}}\in \mathbb{R}^{D^{\text{slow}}\times T^{\text{slow}}\times H'\times W'}$, respectively. Default settings $D^{\text{slow}}=D=2D^{\text{fast}}$ and $T^{\text{fast}}=2T'=2T^{\text{slow}}$ are used for computational tradeoff. Here we set temporal scale to two by considering that there are many low-framerate videos in the VIPL-HR dataset~\cite{niu2019rhythmnet}. Too higher scales would result in the pulse rhythm incompletion/artifacts for high HR values (e.g., >120 bpm). We will investigate more scales for higher framerate videos in the future. Subsequently, the tube tokens from the slow pathway will be forwarded with $N=3N'$ temporal difference periodic transformer blocks while tube tokens from the fast pathway will pass $N'$ temporal difference transformer and $2N'$ temporal difference cross-attention transformer blocks. Specifically, the feature interactions between SlowFast pathways are in two folds: 1) all semantic mid- and high-level features from the slow path are cross-attentive with those from the fast path; and 2) the last mid-level features from two pathways $X^{\text{fast-mid}}_{\text{tube}}$, $X^{\text{slow-mid}}_{\text{tube}}$ are lateral connected and then aggregated for the high-level propagation in the slow pathway. The lateral connection and aggregation can be formulated as
\begin{equation}  \footnotesize
X^{\text{slow-mid}}_{\text{tube}}= \mathrm{Conv2}(\mathrm{Concat}(X^{\text{slow-mid}}_{\text{tube}},  \mathrm{Conv1}(X^{\text{fast-mid}}_{\text{tube}}))),
\end{equation}
where $\text{Conv1}$ is the temporal convolution  with size=3x1x1, stride=2x1x1, padding=1x0x0 while $\text{Conv2}$ denotes the point-wise convolution with $D$ output channel. The lateral connection adaptively transfers the mid-level fine-grained rPPG clues from the Fast pathway to the Slow pathway, and provides complementary temporal details for the Slow pathway to alleviate information loss especially for high-HR scenarios (e.g., after exercise). Finally, the refined high-level rPPG features from fast and slow (upsampled) pathways are concatenated and forwarded the rPPG predictor head with temporally aggregation, upsampling, spatially averaging, and 1D signal $\hat{Y}\in \mathbb{R}^{T}$ projection.

%Sec.~\ref{sec:PhysFormer}
\vspace{0.4em}
\noindent\textbf{Temporal difference multi-head cross- and self-attention.}\quad  
Compared with the slow pathway, the fast pathway has more fine-grained features but conducts inefficient and inaccurate self-attention due to the temporal redundancy/artifacts. To alleviate the weak self-attention issue in the fast pathway, we propose the temporal difference multi-head cross- and self-attention (TD-MHCSA) module, which could be cascaded with ST-FF module to form the temporal difference cross-attention transformer. With TD-MHCSA, the features in the fast pathway can not only be refined by its own self-attention but also the cross-attention between the SlowFast pathways. 

\begin{figure}
%\vspace{-1.0em}
\centering
\includegraphics[scale=0.52]{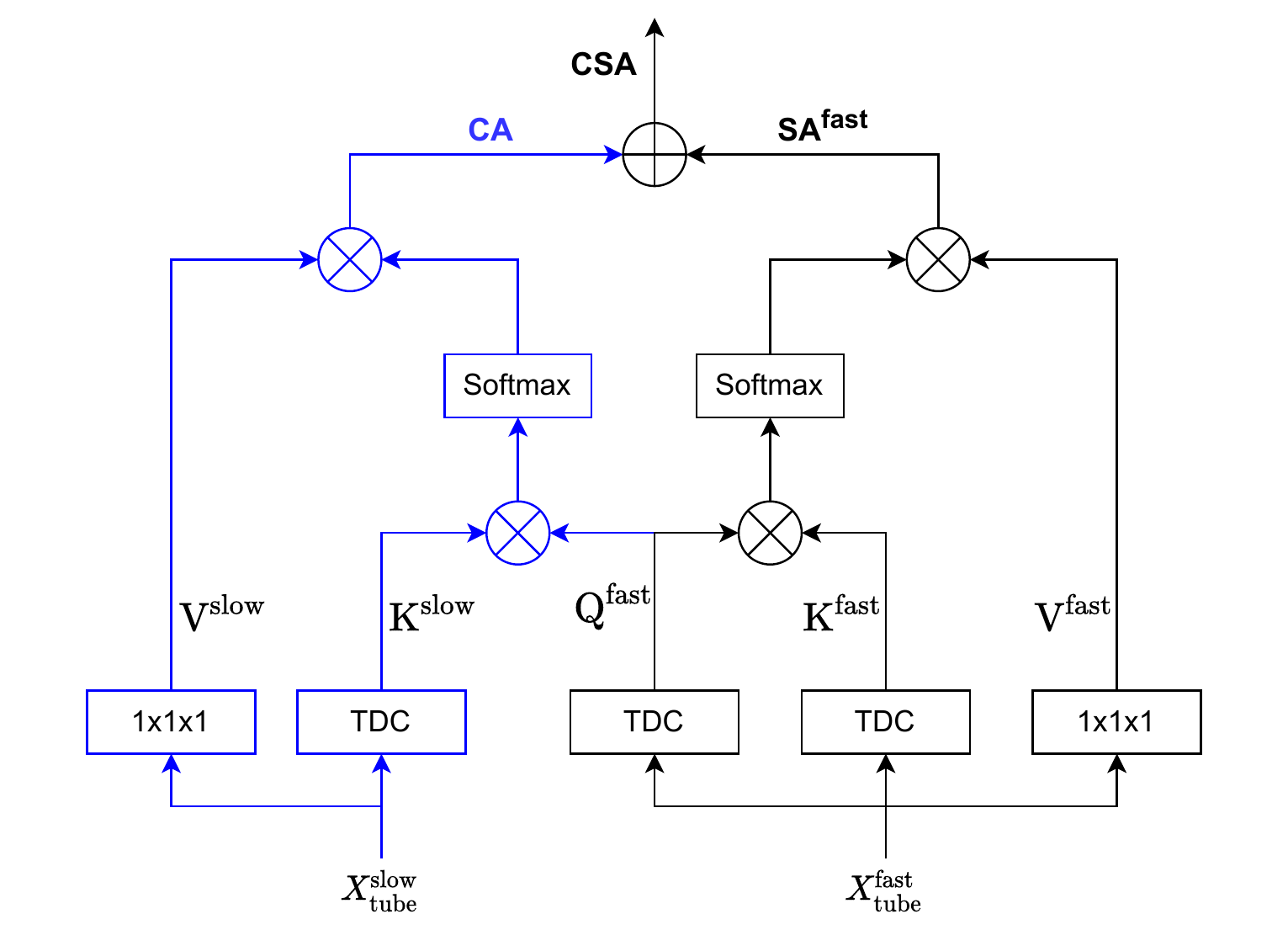}
%\includegraphics[scale=0.34]{KL.pdf}
%\vspace{-1.7em}
  \caption{\small{
  Illustration of the temporal difference multi-head cross- and self-attention (TD-MHCSA) module.}
  }
 
\label{fig:CrossAtten}
%\vspace{-1.2em}
\end{figure}

The structure of the TD-MHCSA is illustrated in Fig.~\ref{fig:CrossAtten}. The features from the fast pathway $X^{\text{fast}}_{\text{tube}}$ are first projected to query and key via
\begin{equation}  \footnotesize
Q^{\text{fast}} = \mathrm{BN}(\mathrm{TDC}(X^{\text{fast}}_{\text{tube}})), K^{\text{fast}}= \mathrm{BN}(\mathrm{TDC}(X^{\text{fast}}_{\text{tube}})). 
\end{equation}
For the value ($V^{\text{fast}}$) projection, point-wise linear projection without BN is utilized. Then $Q^{\text{fast}},K^{\text{fast}},V^{\text{fast}}\in \mathbb{R}^{D^{\text{fast}}\times T^{\text{fast}}\times H'\times W'}$ are flattened into sequence, and separated into $h$ heads ($D^{\text{fast}}_h=D^{\text{fast}}/h$ for each head). For the $i$-th head ($i\leq h$), the self-attention can be formulated
%\vspace{-0.3em}
\begin{equation}
\mathrm{SA^{\text{fast}}_{i}}=\mathrm{Softmax}(Q^{\text{fast}}_{i}{K^{\text{fast}}_{i}}^{T}/\tau)V^{\text{fast}}_{i}.
%\vspace{-0.3em}
\end{equation}
Similarly, the features from the slow pathway $X^{\text{slow}}_{\text{tube}}$ are projected to key $K^{\text{slow}}$ via
%\begin{equation} 
%K^{\text{slow}} = %\mathrm{BN}(\mathrm{TDC}(X^{\text{slow}}_{\text{tube}})), 
%\end{equation}
$\mathrm{BN}(\mathrm{TDC}(X^{\text{slow}}_{\text{tube}}))$ as well as the value ($V^{\text{slow}}$) projection using point-wise linear projection. Then $K^{\text{slow}},V^{\text{slow}}\in \mathbb{R}^{D^{\text{slow}}\times T^{\text{slow}}\times H'\times W'}$ are flattened into sequence, and separated into $h$ heads. For the $i$-th head ($i\leq h$), the cross-attention (CA) can be formulated as
\begin{equation}
\mathrm{CA}_{i}=\mathrm{Softmax}(Q^{\text{fast}}_{i}{K^{\text{slow}}_{i}}^{T}/\tau)V^{\text{slow}}_{i}.
\end{equation} 
Thus, the combined cross- and self-attention (CSA) is formulated as $\mathrm{CSA}_{i}=\mathrm{CA}_{i}+\mathrm{SA_{i}^{\text{fast}}}$.
The output of TD-MHCSA is the concatenation of CSA from all heads and then with a linear projection $U^{\text{fast}}\in \mathbb{R}^{D^{\text{fast}}\times D^{\text{fast}}}$, which is formulated
\begin{equation} \footnotesize
\text{TD-MHCSA} = \mathrm{Concat}(\mathrm{SCA}_{1}; \mathrm{SCA}_{2};...; \mathrm{SCA}_{h})U^{\text{fast}}.
\end{equation}
Finally, residual connection and LN layer would be conducted after TD-MHCSA.

\begin{figure}
%\vspace{-1.0em}
\centering
\includegraphics[scale=0.58]{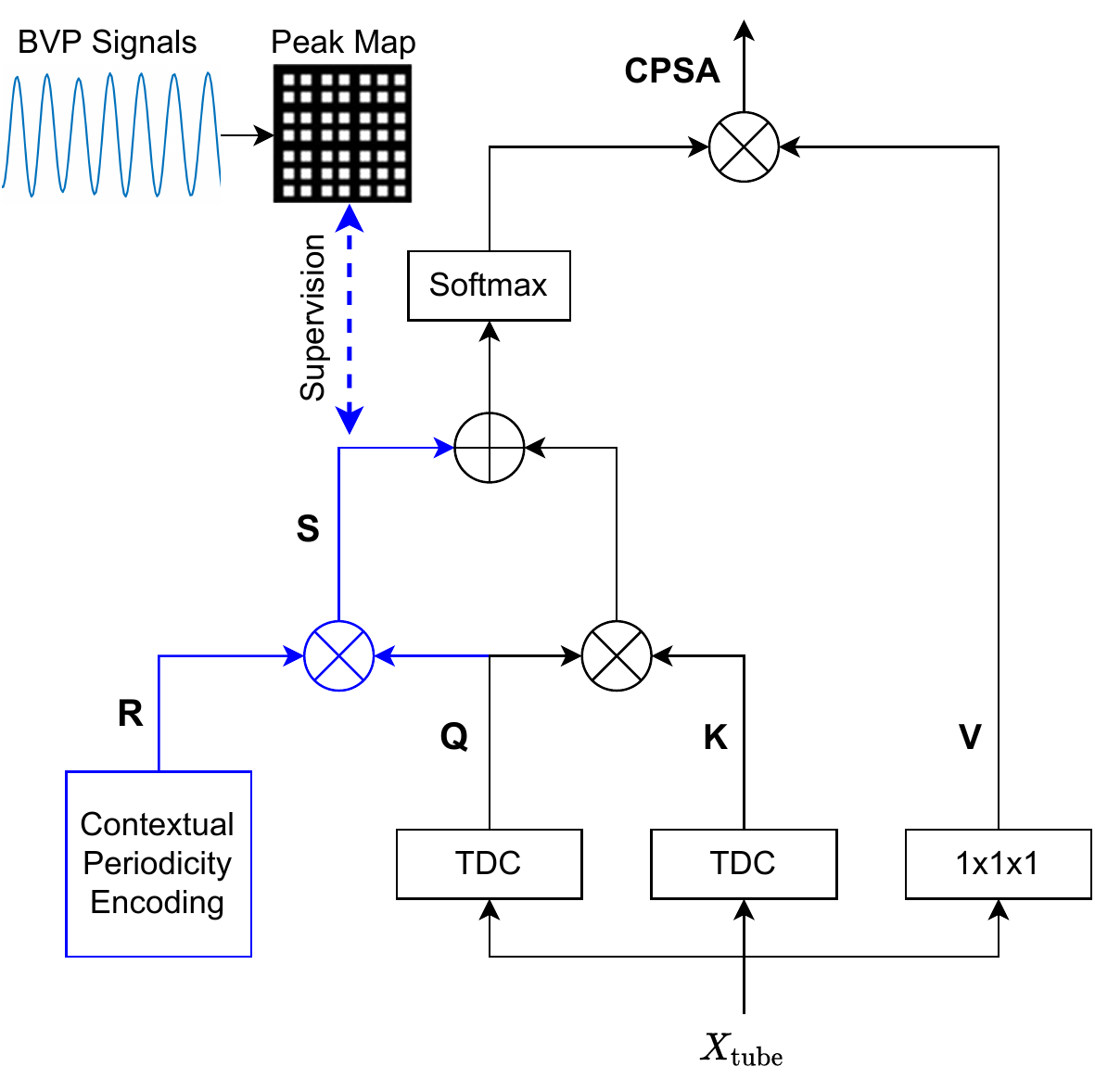}
%\includegraphics[scale=0.34]{KL.pdf}
%\vspace{-1.7em}
  \caption{\small{
  Illustration of the temporal difference multi-head periodic- and self-attention (TD-MHPSA) module.}
  }
 
\label{fig:ContextualAttention}
%\vspace{-1.2em}
\end{figure}

\vspace{0.4em}
\noindent\textbf{Temporal difference multi-head periodic- and self-attention.}\quad
Inspired by the music transformer~\cite{huang2018music} using relative attention~\cite{shaw2018self,wu2021rethinking} to mine richer positional relationship (e.g., periodicity in music signals), we propose the temporal difference multi-head periodic- and self-attention (TD-MHPSA), which extends the TD-MHSA (in Sec.~\ref{sec:PhysFormer}) with learnable rPPG-aware positional contextual periodicity representation. Specifically, as shown in Fig.~\ref{fig:ContextualAttention}, the learnable contextual periodicity encoding $R\in \mathbb{R}^{T'H'W'\times T'H'W' \times D}$ contains the spatio-temporal positional clues, and modulates the query vector $Q$ into the periodic attention $S=QR^{T}$. In consideration of the multi-head $h$ setting, for the $i$-th head, the joint contextual periodicity (CP) and self-attention (SA) can be formulated as
\begin{equation}
\mathrm{CPSA}_{i}=\mathrm{Softmax}((Q_{i}K^{T}_{i}+\lambda \cdot S_{i})/\tau)V_{i},
\end{equation}
where $\lambda$ tradeoffs the CP and SA. Here we follow the memory efficient implementation in~\cite{huang2018music} for $S$ calculation. 

Despite richer positional periodicity clues, the predicted periodic attention $S$ might be easily influenced by some rPPG-unrelated clues (e.g., light changes and dynamic noise). To alleviate this issue, we propose a periodicity constraint to supervise the periodic $S$ representation. As shown in the top left of Fig.~\ref{fig:ContextualAttention}, the approximate peak map $\text{PM}$ can be obtained via 1) first extracting the binary peak signal $P\in \mathbb{R}^{T}$ from the ground truth BVP signal $Y\in \mathbb{R}^{T}$ via
\begin{equation}
P_{t\in T}=\left\{\begin{matrix}
1,\quad if \quad  Y_{t}\in \mathcal{R}_{peak},  \\ 0,\quad if \quad  Y_{t}\notin \mathcal{R}_{peak},
\end{matrix}\right.
\label{eq:peak}
\end{equation}
where $\mathcal{R}_{peak}$ denotes the 1D-region of peak locations; and then 2) calculating the auto-correlation of the peak signal $P$ via $\text{PM}=PP^{T}$. Finally, the periodic-attention loss $\mathcal{L}_{\text{atten}}$ can be calculated with the binary cross-entropy (BCE) loss between the adaptive-spatial-pooled periodic attention maps $S'\in \mathbb{R}^{T'\times T'}$ (from each head and each TD-MHPSA module) and the subsampled binary peak maps $\text{PM'}\in \mathbb{R}^{T'\times T'}$. It can be formulated as
\begin{equation}
\mathcal{L}_{\text{atten}}=\frac{1}{h\times N}\sum_{i\in h,j\in N}\text{BCE}(S',\text{PM'}).
\end{equation}
We also try supervision with $L1$ regression loss instead of BCE loss but with poorer performance.

\vspace{0.4em}
\noindent\textbf{Relationship between PhysFormer and PhysFormer++.}\quad   PhysFormer++ can be treated as an upgraded version of PhysFormer towards excellent performance while with more computational cost. With similar temporal difference transformers, PhysFormer can be seen as a slow-pathway version of PhysFormer++, which is more lightweight and efficient. In contrast, PhysFormer++ is designed based on a dual-pathway SlowFast architecture with complex cross-tempo interactions, which is more robust to head motions and less sensitive to the video framerate, but with heavier computational cost (see Table~\ref{tab:ResultsMMSE} for efficiency analysis).

\subsection{Label Distribution Learning}
\label{sec:distribution}
Similar to the facial age estimation task~\cite{geng2013facial,gao2018age} that faces at close ages look quite similar, facial rPPG signals with close HR values usually have similar periodicity. Inspired by this observation, instead of considering each facial video as an instance with one label (HR), we regard each facial video as an instance associated with a label distribution. The label distribution covers a certain number of class labels, representing the degree that each label describes the instance. Through this way, one facial video can contribute to both targeted HR value and its adjacent HRs.

To consider the similarity information among HR classes during the training stage, we model the rPPG-based HR estimation problem as a specific $L$-class multi-label classification problem, where $L$=139 in our case (each integer HR value within [42, 180] bpm as a class). A label distribution $\mathbf{p}= \left\{ p_1,p_2,...,p_L\right\}\in \mathbb{R}^L$ is assigned to each facial video $X$. It is assumed that each entry of $\mathbf{p}$ is a real value in the range [0,1] such that $\sum_{k=1}^{L}p_k=1$. We consider the Gaussian distribution function, centered at the ground truth HR label $Y_{\text{HR}}$ with the standard deviation $\sigma$, to construct the corresponding label distribution $\mathbf{p}$. 
\begin{equation} 
p_k=\frac{1}{\sqrt{2\pi}\sigma}\exp\left ( -\frac{(k-(Y_{HR}-41))^2}{2\sigma^2} \right ).
\end{equation}
The label distribution loss can be formulated as $\mathcal{L}_{\text{LD}}=\mathrm{KL}(\mathbf{p}, \mathrm{Softmax}(\mathbf{\hat{p}}))$, where divergence measure $\mathrm{KL}(\cdot)$ denotes the Kullback-Leibler (KL) divergence~\cite{gao2017deep}, and $\mathbf{\hat{p}}$ is the power spectral density (PSD) of predicted rPPG signals. 

%We also compare with the case using groundtruth PPG signals based PSD as $\mathbf{p}$ in Table~\ref{tab:ablation3} but with inferior performance.

Please note that the previous work~\cite{niu2017continuous} also considers the distribution learning for HR estimation. However, it is totally different with our work: 1) the motivation in~\cite{niu2017continuous} is to smooth the temporal HR outliers caused by facial movements across continuous video clips, while our work is more generic, aiming at efficient feature learning across adjacent labels under limited-scale training data; 2) the technique used in~\cite{niu2017continuous} is after a post-HR-estimation for the handcrafted rPPG signals, while our work is to design a reasonable supervision signal $\mathcal{L}_{\text{LD}}$ for the  PhysFormer family.

\subsection{Curriculum Learning Guided Dynamic Loss}
\label{sec:dynamic}

Curriculum learning~\cite{bengio2009curriculum}, as a major machine learning regime with philosophy of easy-to-hard curriculum, is utilized to train PhysFormer. In the rPPG measurement task, the supervision signals from temporal domain (e.g., mean square error loss~\cite{chen2018deepphys}, negative Pearson loss~\cite{yu2019remote1,yu2019remote2}) and frequency domain (e.g., cross-entropy loss~\cite{niu2020video,yu2020autohr}, signal-to-noise ratio loss~\cite{vspetlik2018visual}) provide different extents of constraints for model learning. The former one gives signal-trend-level constraints, which is straightforward for model convergence but overfitting after that. In contrast, the latter one with strong constraints on frequency domain enforces the model learning periodic features within target frequency bands, which is hard to converge well due to the realistic rPPG-irrelevant noise. Inspired by the curriculum learning, we propose the dynamic supervision to gradually enlarge the frequency constraints, which alleviates the overfitting issue and benefits the intrinsic rPPG-aware feature learning gradually. Specifically, exponential increment strategy is adopted, and comparison with other dynamic strategies (e.g., linear increment) will be shown in Table~\ref{tab:ablation3}. The dynamic loss $\mathcal{L}_{\text{overall}}$ can be formulated as
%\vspace{-0.1em}
\begin{equation}
\begin{split}
\mathcal{L}_{\text{overall}}&=\underbrace{\alpha\cdot\mathcal{L}_{\text{time}}}_{\text{temporal}}+\underbrace{\beta\cdot(\mathcal{L}_{\text{CE}}+\mathcal{L}_{\text{LD}})}_{\text{frequency}}+\mathcal{L}_{\text{atten}},\\
\beta&=\beta_{0}\cdot(\eta^{({\text{Epoch}}_{\text{current}}-1)/{\text{Epoch}}_{\text{total}}}),
%\vspace{-0.1em}
\end{split}
\end{equation}
where hyperparameters $\alpha$, $\beta_{0}$ and $\eta$ equal to 0.1, 1.0 and 5.0, respectively. Negative Pearson loss~\cite{yu2019remote1,yu2019remote2} and frequency cross-entropy loss~\cite{niu2020video,yu2020autohr} are adopted as $\mathcal{L}_{\text{time}}$ and $\mathcal{L}_{\text{CE}}$, respectively. With the dynamic supervision, PhysFormer and PhysFormer++ could perceive better signal trend at the beginning while such perfect warming up facilitates the gradually stronger frequency knowledge learning later.

\section{Experimental Evaluation}
%%\vspace{-0.1em}
\label{sec:experiment}
In this section, experiments of rPPG-based physiological measurement for three types of physiological signals, i.e., heart rate (HR), heart rate variability (HRV), and respiration frequency (RF), are conducted on four benchmark datasets (VIPL-HR~\cite{niu2019rhythmnet}, MAHNOB-HCI~\cite{soleymani2011multimodal}, MMSE-HR~\cite{tulyakov2016self}, and OBF~\cite{li2018obf}). Besides, comprehensive ablations about PhysFormer and PhysFormer++ are also investigated in the VIPL-HR dataset.

\subsection{Datasets and Performance Metrics}

\begin{figure}
%\vspace{-1.0em}
\centering
\includegraphics[scale=0.5]{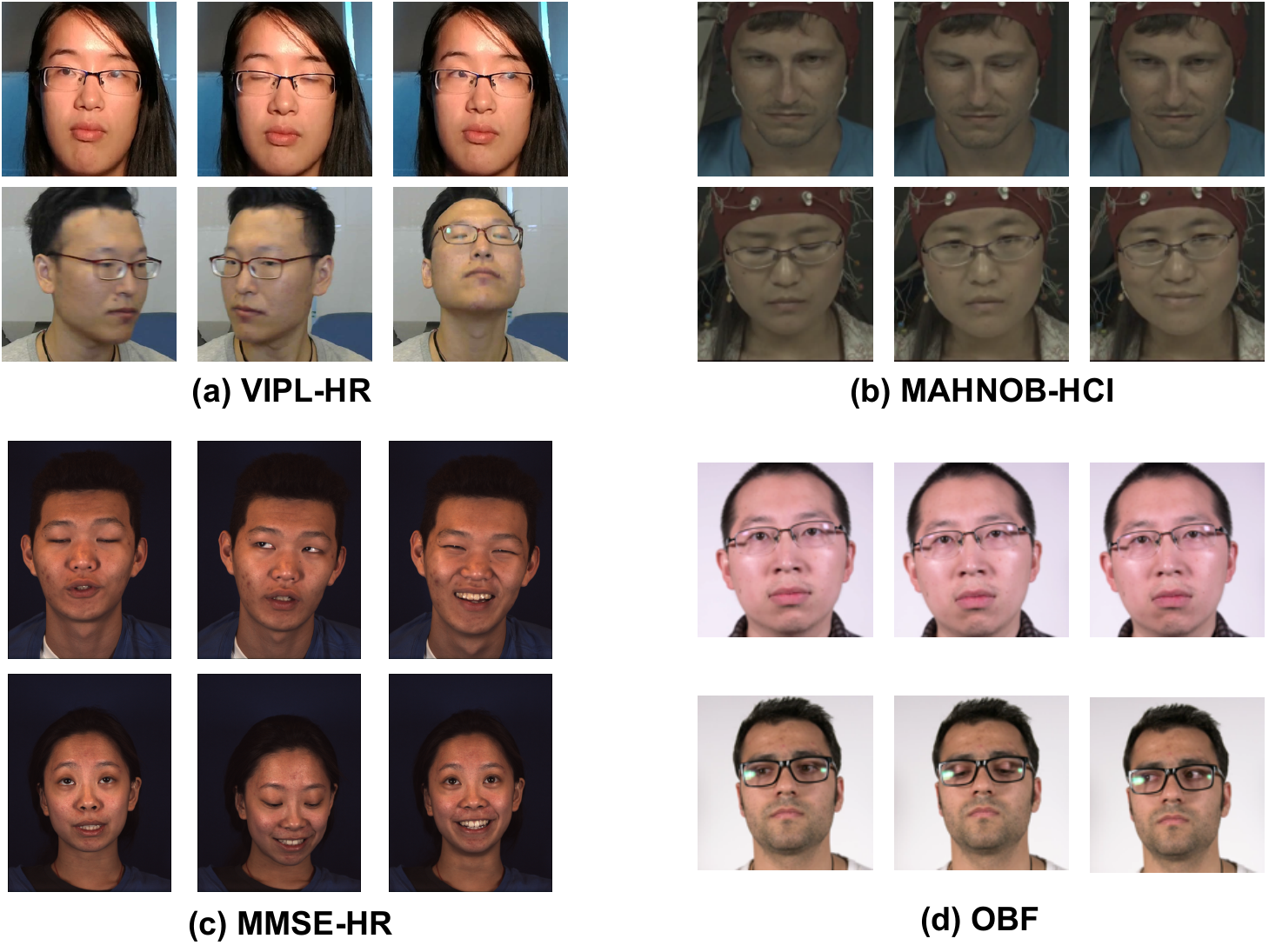}
  \caption{\small{
  Example video frames from datasets (a) VIPL-HR~\cite{niu2019rhythmnet}; (b) MAHNOB-HCI~\cite{soleymani2011multimodal}; (c) MMSE-HR~\cite{tulyakov2016self}; and (d) OBF~\cite{li2018obf}.}
  }
 
\label{fig:dataset}
\end{figure}

\label{sec:dataset} \textbf{VIPL-HR}~\cite{niu2019rhythmnet} is a large-scale dataset for remote physiological measurement under less-constrained scenarios. It contains 2,378 RGB videos of 107 subjects recorded with different head movements, lighting conditions and acquisition devices. \textbf{MAHNOB-HCI}~\cite{soleymani2011multimodal} is one of the most widely used benchmark for remote HR measurement evaluations. It includes 527 facial videos of with 61 fps framerate and 780x580 resolution from 27 subjects. \textbf{MMSE-HR}~\cite{tulyakov2016self} is a dataset including 102 RGB videos from 40 subjects, and the raw resolution of each video is at 1040x1392. \textbf{OBF}~\cite{li2018obf} is a high-quality dataset for remote physiological signal measurement. It contains 200 five-minute-long RGB videos with 60 fps framerate recorded from 100 healthy adults. The example video frames from these four rPPG datasets are illustrated in Fig.~\ref{fig:dataset}.

For MAHNOB-HCI, as there are no available BVP ground truth, we first smooth the sharp ECG signals (with 10-point averaging strategy) into pseudo BVP signals as ground truth. Specifically, to alleviate the incorrect synchronization between videos and ground truth signals in MAHNOB-HCI, OBF, and VIPL-HR datasets, we first extract the coarse green channel signals via averaging the segmented facial skin in each frame. Then, we calculate the cross-correlation between the coarse green rPPG signals and (pseudo) BVP signals, and use the maximum-correlation phase to calibrate/compensate the phase bias. Furthermore, we remove the samples with HR$\textgreater$180 in the VIPL-HR and MMSE-HR datasets because the ground truths in these samples are unreliable due to poor contact of sensors (resulting in very noisy and fluctuated HRs).

In terms of evaluation metrics, average HR estimation task is evaluated on all four datasets while HRV and RF estimation tasks on high-quality OBF~\cite{li2018obf} dataset. Specifically, we follow existing methods~\cite{yu2019remote2,niu2020video,lu2021dual} and report low frequency (LF), high frequency (HF), and LF/HF ratio for HRV and RF estimation. We report the most commonly used performance metrics for evaluation, including the standard deviation (SD), mean absolute error (MAE), root mean square error (RMSE), and Pearson's correlation coefficient ($r$).

\subsection{Implementation Details}
%%\vspace{-0.2em}
\label{sec:Details}

Both PhysFormer and PhysFormer++ are implemented with Pytorch. For each video clip, the MTCNN face detector~\cite{zhang2016joint} is used to crop the enlarged face area at the first frame and fix the region through the following frames. The videos in MAHNOB-HCI and OBF are downsampled to 30 fps for efficiency. The numbers of temporal difference transformer blocks $N$=12, transformer heads $h$=4, channel dimension $D$=96, hidden dimension in ST-FF $D'$=144 are used for PhysFormer while temporal difference coefficient $\theta$=0.7 and attention sparsity $\tau$=2.0 for TD-MHSA. $\lambda$=0.5 is utilized in the TD-MHPSA. The targeted tube size $T_{s}\times H_{s}\times W_{s}$ equals to 4$\times$4$\times$4. For the $\mathcal{R}_{peak}$ calculation in Eq. (\ref{eq:peak}), the function `findpeaks()' in Matlab is used for BVP peak detection, and then the detected peak locations are extended with successive $\pm$3 neighbors. 

In the training stage, we randomly sample RGB face clips with size 160$\times$128$\times$128 ($T\times H\times W$) as model inputs. Random horizontal flipping and temporally up/down-sampling~\cite{yu2020autohr} are used for data augmentation. The PhysFormer is trained with Adam optimizer and the initial learning rate and weight decay are 1e-4 and 5e-5, respectively. We cannot find obvious performance improvement using AdamW optimizer. We train models with 25 epochs with fixed setting $\alpha$=0.1 for temporal loss while exponentially increased parameter $\beta \in [1,5]$ for frequency losses. We set standard deviation $\sigma$=1.0 for label distribution learning. The batch size is 4 on one V100 GPU. In the testing stage, similar to~\cite{niu2019rhythmnet}, we uniformly separate 30-second videos into three short clips with 10 seconds, and then video-level HR is calculated via averaging HRs from three short clips.

\begin{table}[t]\small
\centering
\caption{Intra-dataset testing results on the VIPL-HR dataset. The symbols $\blacktriangle$, $\blacklozenge$ and $\star$ denote traditional, non-end-to- end learning based and end-to-end learning based methods, respectively. Best results are marked in \textbf{bold} and second best in \underline{underline}.} \label{tab:ResultsVIPL}

%\vspace{-0.8em}

\resizebox{0.48\textwidth}{!} {\begin{tabular}{l c c c c} 
 \toprule%\toprule[1pt]
 Method & \begin{tabular}[c]{@{}c@{}}SD $\downarrow$ \\(bpm)\end{tabular} & \begin{tabular}[c]{@{}c@{}}MAE $\downarrow$ \\(bpm)\end{tabular}  & \begin{tabular}[c]{@{}c@{}}RMSE $\downarrow$ \\(bpm)\end{tabular}  & $r$ $\uparrow$\\
 %\hline
 \midrule
 Tulyakov2016~\cite{tulyakov2016self}$\blacktriangle$ & 18.0 & 15.9 & 21.0 &  0.11\\
 POS~\cite{wang2017algorithmic}$\blacktriangle$ & 15.3 & 11.5 & 17.2 & 0.30 \\
 CHROM~\cite{de2013robust}$\blacktriangle$ & 15.1 & 11.4 & 16.9 & 0.28 \\ 
 \midrule
 RhythmNet~\cite{niu2019rhythmnet}$\blacklozenge$ & 8.11 & 5.30 & 8.14 &  0.76\\
 ST-Attention~\cite{niu2019robust}$\blacklozenge$ & 7.99 & 5.40 & 7.99 &  0.66\\
 NAS-HR~\cite{lu2021hr}$\blacklozenge$ & 8.10 & 5.12 & 8.01 & 0.79\\
 CVD~\cite{niu2020video}$\blacklozenge$ & 7.92 & 5.02 &7.97 &  0.79\\
  Dual-GAN~\cite{lu2021dual}$\blacklozenge$ & \textbf{7.63} & \underline{4.93} & \underline{7.68} &  \textbf{0.81}\\
 \midrule

 I3D~\cite{carreira2017quo}$\star$ & 15.9  & 12.0 & 15.9 & 0.07\\
  PhysNet~\cite{yu2019remote1}$\star$ & 14.9 & 10.8 & 14.8 & 0.20 \\
 DeepPhys~\cite{chen2018deepphys}$\star$ & 13.6 & 11.0  & 13.8 & 0.11 \\ 
  VideoTransformer~\cite{revanur2022instantaneous}$\star$ & 13.5 & 10.4  & 13.2 & 0.16 \\ 
AutoHR~\cite{yu2020autohr}$\star$ & 8.48
 & 5.68 & 8.68 & 0.72\\

 \textbf{PhysFormer (Ours)$\star$} & 7.74 & 4.97 & 7.79 & 0.78\\

 \textbf{PhysFormer++ (Ours)$\star$} & \underline{7.65}  &  \textbf{4.88}  & \textbf{7.62}  & \underline{0.80} \\

  %\hline
 %\hline
 \bottomrule[0.8pt]
 \end{tabular}}
 %\vspace{-1.0em}
\end{table}

\subsection{Intra-dataset Testing}
In this subsection, two datasets (VIPL-HR and MAHNOB-HCI) are used for intra-dataset testing on HR estimation while the OBF dataset is used for intra-dataset HR, HRV and RF estimation.

\vspace{0.4em}
\noindent\textbf{HR estimation on VIPL-HR.} \quad   Here we follow~\cite{niu2019rhythmnet} and use a
subject-exclusive 5-fold cross-validation protocol on VIPL-HR. As shown in Table~\ref{tab:ResultsVIPL}, all three traditional methods (Tulyakov2016~\cite{tulyakov2016self}, POS~\cite{wang2017algorithmic} and CHROM~\cite{de2013robust}) perform poorly due to the complex scenarios (e.g., large head movement and various illumination) in the VIPL-HR dataset. In terms of deep learning based methods, the existing end-to-end learning based methods (e.g., PhysNet~\cite{yu2019remote1}, DeepPhys~\cite{chen2018deepphys}, and AutoHR~\cite{yu2020autohr}) predict less reliable HR values with larger RMSE compared with non-end-to-end learning approaches (e.g., RhythmNet~\cite{niu2019rhythmnet}, ST-Attention~\cite{niu2019robust}, NAS-HR~\cite{lu2021hr}, CVD~\cite{niu2020video}, and Dual-GAN~\cite{lu2021dual}). Such the large performance margin might be caused by the coarse and overfitted rPPG features extracted from the end-to-end models. In contrast, all five non-end-to-end methods first extract fine-grained signal maps from multiple facial ROIs, and then more dedicated rPPG clues would be extracted via the cascaded models. Without strict and heavy preprocessing procedure in~\cite{niu2019rhythmnet,niu2019robust,lu2021hr,niu2020video,lu2021dual}, the proposed PhysFormer and PhysFormer++ can be trained from scratch on facial videos directly, and achieve better or on par performance with state-of-the-art non-end-to-end learning based method Dual-GAN~\cite{lu2021dual}. It indicates that PhysFormer and PhysFormer++ are able to learn the intrinsic and periodic rPPG-aware features automatically. It can be seen from Table~\ref{tab:ResultsVIPL} that the proposed PhysFormer family outperforms the VideoTransformer~\cite{revanur2022instantaneous} by a large margin, indicating the importance of local and global spatio-temporal physiological propagation.

\begin{table}[t]\small
\centering
\caption{Intra-dataset results on the  MAHNOB-HCI dataset.} \label{tab:ResultsMAHNOB}

%\vspace{-1.0em}

\resizebox{0.48\textwidth}{!} {\begin{tabular}{l c c c c} 
 %\hline\hline
 \toprule%\toprule[1pt]
 Method & \begin{tabular}[c]{@{}c@{}}SD $\downarrow$ \\(bpm)\end{tabular} & \begin{tabular}[c]{@{}c@{}}MAE $\downarrow$ \\(bpm)\end{tabular}  & \begin{tabular}[c]{@{}c@{}}RMSE $\downarrow$ \\(bpm)\end{tabular}  & $r$ $\uparrow$\\
 %\hline
 \midrule
 Poh2010~\cite{poh2010advancements}$\blacktriangle$ & 13.5 & - & 13.6 & 0.36 \\ 
 CHROM~\cite{de2013robust}$\blacktriangle$ & - & 13.49 & 22.36 & 0.21 \\
 Li2014~\cite{li2014remote}$\blacktriangle$ & 6.88 & - & 7.62 & 0.81\\
 Tulyakov2016~\cite{tulyakov2016self}$\blacktriangle$ & 5.81 & 4.96 & 6.23 & 0.83\\
  \midrule
 SynRhythm~\cite{niu2018synrhythm}$\blacklozenge$ & 10.88 & - & 11.08 & - \\ 
 RhythmNet~\cite{niu2019rhythmnet}$\blacklozenge$ & 3.99 & - & 3.99  & \textbf{0.87} \\ 
 \midrule
 HR-CNN~\cite{vspetlik2018visual}$\star$ & - & 7.25 & 9.24 & 0.51 \\
  rPPGNet~\cite{yu2019remote2}$\star$ & 7.82 & 5.51 & 7.82 & 0.78 \\
  DeepPhys~\cite{chen2018deepphys}$\star$ & - & 4.57 & - & -\\
 AutoHR~\cite{yu2020autohr}$\star$ & 4.73 & 3.78 & 5.10 & \underline{0.86}\\
 Meta-rPPG~\cite{lee2020meta}$\star$ & 4.9 & \textbf{3.01} & \textbf{3.68} & 0.85\\
 
 \textbf{PhysFormer (Ours)$\star$} & \textbf{3.87} & 3.25 & 3.97 & \textbf{0.87}\\
  
 \textbf{PhysFormer++ (Ours)$\star$} & \underline{3.90} & \underline{3.23} & \underline{3.88} & \textbf{0.87}\\
 
  %\hline
 %\hline
 \bottomrule[0.8pt]
 \end{tabular}}
 %\vspace{-0.8em}
\end{table}

\begin{table*}[t]\footnotesize
\begin{center}
\caption{Performance comparison of HR and RF measurement as well as HRV analysis on the OBF dataset. }

%\vspace{-1.0em}

\label{tab:OBFrPPGNet}
\centering
\resizebox{0.91\textwidth}{!} {\begin{tabular}{p{4.5cm}  p{0.5cm} p{0.6cm}  p{0.5cm} p{0.6cm}   p{0.5cm} p{0.6cm}  p{0.5cm} p{0.6cm} p{0.5cm} p{0.6cm}}

%\toprule%\toprule[1pt]
\toprule%\toprule[1pt]
& \multicolumn{2}{c}{HR(bpm)} &  \multicolumn{2}{c}{RF(Hz)} &  \multicolumn{2}{c}{LF(u.n)} &  \multicolumn{2}{c}{HF(u.n)} &  \multicolumn{2}{c}{LF/HF}\\
  
\cmidrule(lr){2-3} \cmidrule(lr){4-5}
\cmidrule(lr){6-7} \cmidrule(lr){8-9}
\cmidrule(lr){10-11}

  \multicolumn{1}{c}{Method} &  RMSE & \quad$r$  &   RMSE & \quad$r$  &  RMSE & \quad$r$  & RMSE & \quad$r$  &   RMSE & \quad$r$\\

 %\hline
 \midrule
 ROI\_green~\cite{li2018obf}$\blacktriangle$    & 2.162 & 0.99   &  0.084  & 0.321  & 0.24 & 0.573 & 0.24 & 0.573 & 0.832 & 0.571\\
 
 % CHROM~\cite{de2013robust}
 CHROM~\cite{de2013robust}$\blacktriangle$   & 2.733 & 0.98    & 0.081  & 0.224  & 0.206 & 0.524 & 0.206 & 0.524 &  0.863 & 0.459\\
 
 POS~\cite{wang2017algorithmic}$\blacktriangle$   %& 1.899 & 1.906 & 0.991   & 0.07 & 0.07  & 0.44  & 0.155  & 0.158 & 0.727 & 0.155  & 0.158 & 0.727 & 0.663  & 0.679 & 0.687\\
 & 1.906 & 0.991    & 0.07  & 0.44    & 0.158 & 0.727  & 0.158 & 0.727  & 0.679 & 0.687\\
  \midrule

 CVD~\cite{niu2020video}$\blacklozenge$  &  1.26 & 0.996  & 0.058  & 0.606 & 0.09 & 0.914 &  0.09 & 0.914 &  0.453 & 0.877\\

 \midrule

 rPPGNet~\cite{yu2019remote2}$\star$  & 1.8 & 0.992 & 0.064   & 0.53 & 0.135  &  0.804 & 0.135  & 0.804 & 0.589  & 0.773\\
 
  \textbf{PhysFormer (Ours)}$\star$  &   \underline{0.804}  & \textbf{0.998} & \underline{0.054}   & \underline{0.661}  & \underline{0.086} & \underline{0.912} & \underline{0.086} & \underline{0.912}  & \underline{0.39} & \underline{0.896}\\

   \textbf{PhysFormer++ (Ours)}$\star$  & \textbf{0.765}  & \textbf{0.998} & \textbf{0.052}   & \textbf{0.686}  & \textbf{0.083} & \textbf{0.921} & \textbf{0.083} & \textbf{0.921}  & \textbf{0.368} & \textbf{0.908}\\
 
\bottomrule[0.8pt]
\end{tabular}}
\end{center}
%\vspace{-2.2em}
\end{table*}

\begin{figure*}
%\vspace{-1.0em}
\centering
\includegraphics[scale=0.46]{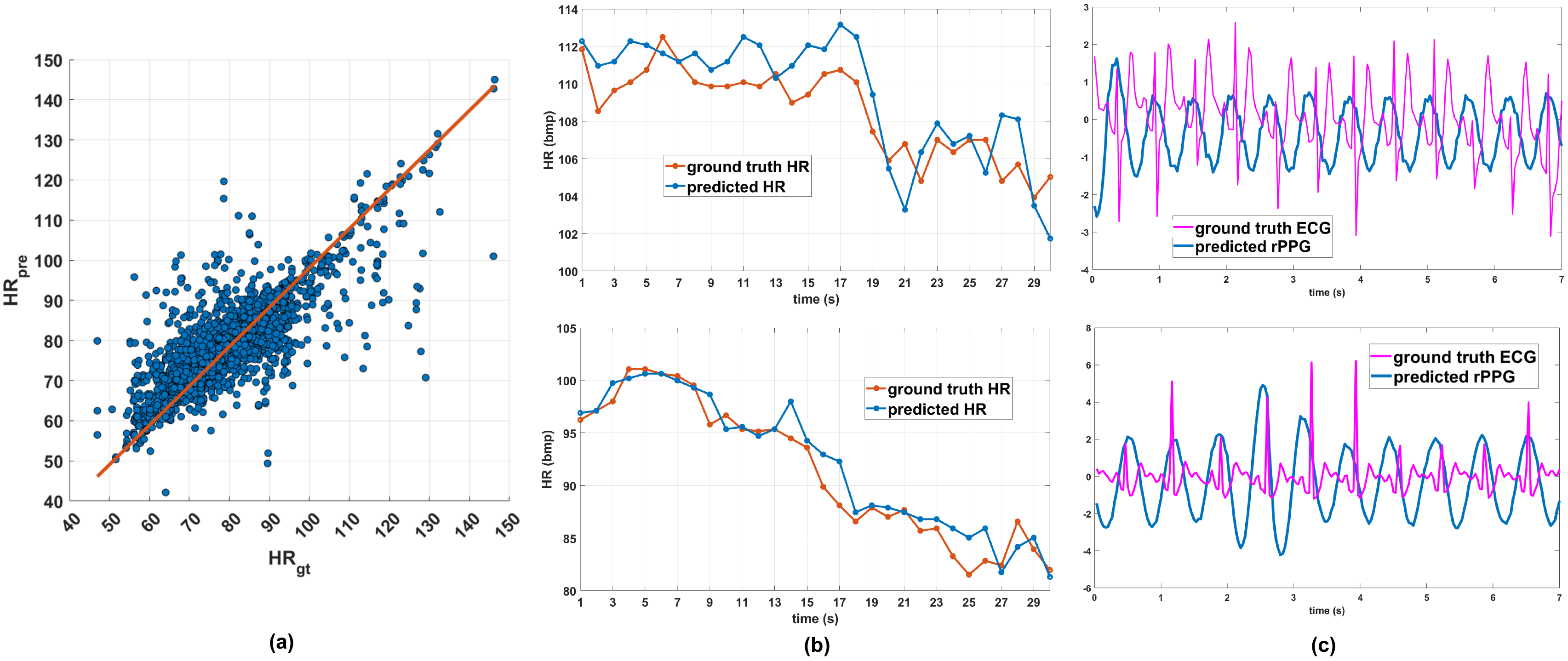}
%\vspace{-1.7em}
\caption{\small{ (a) The scatter plot of the ground truth $\text{HR}_{\text{gt}}$ and the predicted $\text{HR}_{\text{pre}}$ via PhysFormer++ of all the face videos on VIPL-HR dataset. (b) Two examples of the short-time HR estimation from PhysFormer++ for face videos with significantly decreased HR. (c) Two example curves of the predicted rPPG signals from PhysFormer++ and the ground truth ECG signals used to calculate the HRV features.}
}
 
\label{fig:visualization_IJCV_all}
%\vspace{-1.2em}
\end{figure*}

In order to further check the
correlations between the predicted HRs and the ground-truth HRs, we plot the
HR estimation results against the ground truths in Fig.~\ref{fig:visualization_IJCV_all}(a). From the figure we
can see that the predicted HRs from PhysFormer++ and the ground-truth HRs are well correlated in a wide range of HR from 47 bpm to 147 bpm.

\vspace{0.4em}
\noindent\textbf{HR estimation on MAHNOB-HCI.} \quad   
For the HR estimation tasks on MAHNOB-HCI, similar to~\cite{yu2019remote2}, subject-independent 9-fold cross-validation protocol is adopted. In consideration of the convergence difficulty due to the low illumination and high compression videos in MAHNOB-HCI, we finetune the VIPL-HR pretrained models on MAHNOB-HCI for further 15 epochs. The HR estimation results are shown in Table~\ref{tab:ResultsMAHNOB}. The proposed PhysFormer and PhysFormer++ achieves the lowest SD (3.87 bpm) and highest $r$ (0.87) among the traditional, non-end-to-end learning, and end-to-end learning methods, which indicates the reliability of the learned rPPG features from PhysFormer family under sufficient supervision. Our performance is on par with the latest end-to-end learning method Meta-rPPG~\cite{lee2020meta} without transductive adaptation from target frames.

\begin{table}[t]\small
\centering
\caption{Cross-dataset results on the MMSE-HR dataset.} \label{tab:ResultsMMSE}

%\vspace{-1.0em}

\resizebox{0.47\textwidth}{!} {\begin{tabular}{l c c c c} 
 %\hline\hline
 \toprule%\toprule[1pt]
 Method & \begin{tabular}[c]{@{}c@{}}SD $\downarrow$ \\(bpm)\end{tabular}& \begin{tabular}[c]{@{}c@{}}MAE $\downarrow$ \\(bpm)\end{tabular} & \begin{tabular}[c]{@{}c@{}}RMSE $\downarrow$ \\(bpm)\end{tabular}  & $r$ $\uparrow$\\
 %\hline
 \midrule
 Li2014~\cite{li2014remote}$\blacktriangle$ & 20.02 & - & 19.95 & 0.38\\
 CHROM~\cite{de2013robust}$\blacktriangle$ & 14.08 & - & 13.97 & 0.55 \\
 Tulyakov2016~\cite{tulyakov2016self}$\blacktriangle$ & 12.24 & - & 11.37 & 0.71\\
 \midrule
 ST-Attention~\cite{niu2019robust}$\blacklozenge$ & 9.66 & - & 10.10 & 0.64 \\ 
 RhythmNet~\cite{niu2019rhythmnet}$\blacklozenge$ & 6.98 & - & 7.33 & 0.78 \\ 
  CVD~\cite{niu2020video}$\blacklozenge$ & 6.06  & - &  6.04 & 0.84 \\ 
 \midrule
 PhysNet~\cite{yu2019remote1}$\star$ & 12.76 & - & 13.25 & 0.44 \\
 TS-CAN~\cite{liu2020multi}$\star$ & - & 3.85 & 7.21 & 0.86 \\
AutoHR~\cite{yu2020autohr}$\star$ & 5.71 & - & 5.87 & 0.89\\
EfficientPhys-C~\cite{liu2021efficientphys}$\star$ & - &  2.91 & 5.43 & \underline{0.92}\\
EfficientPhys-T1~\cite{liu2021efficientphys}$\star$ & - & 3.48 & 7.21 & 0.86\\
\textbf{PhysFormer (Ours)$\star$} & \underline{5.22} & \underline{2.84} & \underline{5.36} & \underline{0.92}\\

\textbf{PhysFormer++ (Ours)$\star$} & \textbf{5.09} & \textbf{2.71} & \textbf{5.15} & \textbf{0.93}\\

 \bottomrule[0.8pt]
 \end{tabular}}
 %\vspace{-0.8em}
\end{table}

\vspace{0.4em}
\noindent\textbf{HR, HRV and RF estimation on OBF.} \quad  Besides HR estimation, we also conduct experiments for three types of physiological signals, i.e., HR, RF, and HRV measurement on the OBF~\cite{li2018obf} dataset. Following~\cite{yu2019remote2,niu2020video}, we use a 10-fold subject-exclusive protocol for all experiments. All the results are shown in Table~\ref{tab:OBFrPPGNet}. It is clear that the proposed PhysFormer and PhysFormer++ outperform the existing state-of-the-art traditional (ROI\_green~\cite{li2018obf}, CHROM~\cite{de2013robust}, POS~\cite{wang2017algorithmic}) and end-to-end learning (rPPGNet~\cite{yu2019remote2}) methods by a large margin on all evaluation metrics for HR, RF and all HRV features. The proposed PhysFormer and PhysFormer++ give more accurate estimation in terms of HR, RF, and LF/HF compared with the preprocessed signal map based non-end-to-end learning method CVD~\cite{niu2020video}. These results indicate that PhysFormer family could not only handle the average HR estimation task but also give a promising prediction of the rPPG signal for RF measurement and HRV analysis, which shows its potential in many healthcare applications. 

We also check the short-time HR estimation performance of the after exercising scenario on the OBF, in which the subject’s HR decreases rapidly. Two examples are given in Fig.~\ref{fig:visualization_IJCV_all}(b). It can be seen that PhysFormer++ could follow the trend of HR changes well, which indicates the proposed model is robust in the significant HR changing scenarios. We further check the predicted rPPG signals of the PhysFormer++ from these two examples in Fig.~\ref{fig:visualization_IJCV_all}(c). From the results, we can see that the proposed method could give an accurate prediction of the interbeat intervals (IBIs), thus can give a robust estimation of RF and HRV features.

%\vspace{-0.3em}
\subsection{Cross-dataset Testing}
%\vspace{-0.3em}
Besides of the intra-dataset testings on the VIPL-HR, MAHNOB-HCI, and OBF datasets, we also conduct cross-dataset testings on MMSE-HR~\cite{tulyakov2016self} following the protocol of~\cite{niu2019rhythmnet}. The models trained on VIPL-HR are directly tested on MMSE-HR. All the results of the proposed PhysFormer family and the state-of-the-art methods are shown in Table~\ref{tab:ResultsMMSE}. It is clear that PhysFormer and PhysFormer++ generalize well in unseen domains (e.g., skin tone and lighting conditions). It is worth noting that PhysFormer++ achieves the lowest SD (5.09 bpm), MAE (2.71 bpm), RMSE (5.15 bpm) as well as the highest $r$ (0.93) among the traditional, non-end-to-end learning and end-to-end learning based methods, indicating 1) the predicted HRs are highly correlated with the ground truth HRs, and 2) the model learns domain-invariant intrinsic rPPG-aware features. Compared with the spatio-temporal transformer based EfficientPhys-T1~\cite{liu2021efficientphys}, our proposed PhysFormer and PhysFormer++ are able to predict more accurate physiological signals, which indicates the effectiveness of the long-range spatio-temporal attention.

\subsection{Ablation Study}
\label{sec:ablation}
Here We provide the results of ablation studies for HR estimation on the \textbf{Fold-1} of the VIPL-HR~\cite{niu2019rhythmnet} dataset. Specifically, we first evaluate the impacts of architecture configurations for PhysFormer in terms of ‘Tube Tokenization’, ‘TD-MHSA’ and ‘ST-FF’. Then based on the optimal configuration of PhysFormer, the impacts of architecture configurations of PhysFormer++ with ‘TD-MHPSA’ and ‘SlowFast architecture’ will be studied. Finally, we study the transformer configurations (‘$\theta$ in TDC’ and ‘layer/head numbers’) and the training receipts (‘label distribution learning’ and ‘dynamic supervision’) for the whole PhysFormer family (i.e., PhysFormer and PhyFormer++).  

\vspace{0.4em}
\noindent\textbf{Impact of tube tokenization in PhysFormer}.\quad   In the default setting of PhysFormer, a shallow stem cascaded with a tube tokenization is used. In this ablation, we consider other four tokenization configurations with or w/o stem. It can be seen from the first row in Table~\ref{tab:tube} that the stem helps the PhysFormer see better~\cite{xiao2021early}, and the RMSE increases dramatically (+3.06 bpm) when w/o the stem. Then we investigate the impacts of the spatial and temporal domains in tube tokenization. It is clear that the result in the fourth row with full spatial projection is quite poor (RMSE=10.61 bpm), indicating the necessity of the spatial attention. In contrast, tokenization with smaller tempos (e.g., [2x4x4]) or spatial inputs (e.g., 160x96x96) reduces performance slightly. Based on the observed results, tokenizations with [4x4x4] and [2x4x4] are adopted for the defaulted setting of slow and fast pathway in PhysFormer++, respectively.

\begin{table}\small
\centering

\caption{Ablation of Tube Tokenization of PhysFormer. The three dimensions in tensors indicate length$\times$ height$\times$width.}

%\vspace{-0.9em}

%%\vspace{-0.8em}
\scalebox{0.76}{\begin{tabular}{l|c | c |c}
\toprule%\toprule[1pt]
Inputs & \begin{tabular}[c]{@{}c@{}} [Stem] \\ Feature Size\end{tabular} & \begin{tabular}[c]{@{}c@{}} [Tube Size] \\ Token Numbers\end{tabular} & \begin{tabular}[c]{@{}c@{}}RMSE $\downarrow$ \\(bpm)\end{tabular} \\
\hline
$160 \times 128 \times 128$ & \begin{tabular}[c]{@{}c@{}} [$\bigtimes$] \\ $160 \times 128 \times 128$ \end{tabular} & \begin{tabular}[c]{@{}c@{}} $[4 \times 32 \times 32]$ \\ $40 \times 4 \times 4$ \end{tabular} & 10.62 \\
 \hline
$160 \times 128 \times 128$ & \begin{tabular}[c]{@{}c@{}} [$\surd$] \\ $160 \times 16 \times 16$ \end{tabular} & \begin{tabular}[c]{@{}c@{}} $[4 \times 4 \times 4]$ \\ $40 \times 4 \times 4$ \end{tabular} & \textbf{7.56} \\
 \hline
$160 \times 96 \times 96$ & \begin{tabular}[c]{@{}c@{}} [$\surd$] \\ $160 \times 12 \times 12$ \end{tabular} & \begin{tabular}[c]{@{}c@{}} $[4 \times 4 \times 4]$ \\ $40 \times 3 \times 3$ \end{tabular} & 8.03 \\
 \hline
$160 \times 128 \times 128$ & \begin{tabular}[c]{@{}c@{}} [$\surd$] \\ $160 \times 16 \times 16$ \end{tabular} & \begin{tabular}[c]{@{}c@{}} $[4 \times 16 \times 16]$ \\ $40 \times 1 \times 1$ \end{tabular} & 10.61 \\
 \hline
 $160 \times 128 \times 128$ & \begin{tabular}[c]{@{}c@{}} [$\surd$] \\ $160 \times 16 \times 16$ \end{tabular} & \begin{tabular}[c]{@{}c@{}} $[2 \times 4 \times 4]$ \\ $80 \times 4 \times 4$ \end{tabular} & 7.81 \\
 
\bottomrule[0.8pt]

\end{tabular}}

\label{tab:tube}
%\vspace{-0.5em}
\end{table}

\begin{table}
\centering

\caption{Ablation of TD-MHSA and ST-FF in PhysFormer. }

%\vspace{-0.9em}

%%\vspace{-0.8em}
\scalebox{0.84}{\begin{tabular}{l|c | c |c}
\toprule%\toprule[1pt]
MHSA & $\tau$ & Feed-forward & RMSE (bpm) $\downarrow$\\
\hline
- & - & ST-FF & 9.81 \\
TD-MHSA & $\sqrt{D_{h}}\approx$ 4.9 & ST-FF & 9.51 \\
TD-MHSA & 2.0 & ST-FF & \textbf{7.56} \\
vanilla MHSA & 2.0 & ST-FF & 10.43 \\
TD-MHSA & 2.0 & vanilla FF & 8.27 \\
 
\bottomrule[0.8pt]

\end{tabular}}

\label{tab:ablation2}
%\vspace{-0.5em}
\end{table}

\begin{table}
\centering

\caption{Ablation of TD-MHPSA for the single pathway configuration in PhysFormer++. }

%\vspace{-0.9em}

%%\vspace{-0.8em}
\scalebox{0.84}{\begin{tabular}{l|c | c |c}
\toprule%\toprule[1pt]
Pathway & MHSA & $\mathcal{L}_{\text{atten}}$ & RMSE (bpm) $\downarrow$\\
\hline

Slow & TD-MHSA & - & 7.56 \\
Slow & TD-MHPSA & - & 7.69 \\
Slow & TD-MHPSA & $\surd$ & \textbf{7.43} \\

\hline

Fast & TD-MHSA & - & 7.81 \\
Fast & TD-MHPSA & - & 8.12 \\
Fast & TD-MHPSA & $\surd$ & 7.85 \\

\bottomrule[0.8pt]

\end{tabular}}

\label{tab:MHPSA}
%\vspace{-0.5em}
\end{table}

\begin{table}
\centering

\caption{Ablation of SlowFast two-pathway based architecture in PhysFormer++. }

%\vspace{-0.9em}

%%\vspace{-0.8em}
\scalebox{0.8}{\begin{tabular}{l|c | c |c}
\toprule%\toprule[1pt]
TD-MHPSA & Lateral Connect & TD-MHCSA & RMSE\\
\hline

- & - & - & 7.78 \\
Slow Pathway & - & - & 7.58 \\
Slow Pathway & High-level & - & 7.34 \\
Slow Pathway & Mid\&High-level & - & 7.38 \\
Slow Pathway & High-level & High-level & 7.28 \\
Slow Pathway & High-level & Mid\&High-level & \textbf{7.16} \\
Slow Pathway & High-level & Low\&Mid\&High-level & 7.24 \\

\bottomrule[0.8pt]

\end{tabular}}

\label{tab:Slowfast}
%\vspace{-0.5em}
\end{table}

\vspace{0.4em}
\noindent\textbf{Impact of TD-MHSA and ST-FF in PhysFormer}.\quad   
As shown in Table~\ref{tab:ablation2}, both the TD-MHSA and ST-FF play vital roles in PhysFormer. The result in the first row shows that the performance degrades sharply without spatio-temporal attention. Moreover, it can be seen from the last two rows that without TD-MHSA/ST-FF, PhysFormer with vanilla MHSA/FF obtains 10.43/8.27 bpm RMSE. Thus, we can draw the conclusion that the key element ‘vanilla MHSA’ in transformer cannot provide rPPG performance gain although it captures the long-term global spatio-temporal physiological features. In contrast, the proposed ‘TD-MHSA’ benefits the rPPG measurement via local spatio-temporal physiological clue guided long-term global spatio-temporal physiological aggregation. One important finding in this research is that, the temperature $\tau$ influences the MHSA a lot. When the $\tau=\sqrt{D_{h}}$ like previous ViT~\cite{dosovitskiy2020image,arnab2021vivit}, the predicted rPPG signals are unsatisfied (RMSE=9.51 bpm). Regularizing the $\tau$ with smaller value enforces sparser spatio-temporal attention, which is effective for the quasi-periodic rPPG task.

\vspace{0.4em}
\noindent\textbf{Impact of TD-MHPSA for different pathway in PhysFormer++}.\quad 
Based on the TD-MHSA in PhysFormer, the PhysFormer++ further extends the slow pathway with the more periodic TD-MHPSA modules. Table~\ref{tab:MHPSA} shows the results of the TD-MHPSA for single pathway configuration. It is interesting to find that compared with TD-MHSA, the performance even drop for both slow and fast pathways when assembling with TD-MHPSA but without explicit attention supervision $\mathcal{L}_{\text{atten}}$. When training TD-MHPSA with $\mathcal{L}_{\text{atten}}$, the RMSE is decreased by 0.26 and 0.27 bpm for the slow and fast pathway, respectively. It indicates the importance of explicit rPPG-aware periodicity supervision. Some visualizations with and without $\mathcal{L}_{\text{atten}}$ can be found in Sec.~\ref{sec:Analysis}. 

From the results in Table~\ref{tab:MHPSA} we can see that the TD-MHPSA with $\mathcal{L}_{\text{atten}}$ benefits the periodic rPPG clue mining in the slow pathway while limited effects for the fast pathway. It may be because the attention loss calculated from the periodic maps with huger temporal resolution in the fast pathway is inefficient to back-propagate the rPPG-aware information. Thus, we only apply the TD-MHPSA in the slow pathway as the defaulted setting for PhysFormer++.

\vspace{0.4em}
\noindent\textbf{Impact of the SlowFast architecture in PhysFormer++}.\quad 
Table~\ref{tab:Slowfast} illustrates the ablations of SlowFast two-pathway based architecture in PhysFormer++. From the results of the first two rows we can see that such SlowFast rPPG models even achieve inferior performance (7.78/7.58 vs. 7.56 bpm RMSE) compared with single pathway based PhysFormer. The unsatisfied results might be caused by the lack of efficient rPPG feature interaction between two pathways. We also conduct experiments with lateral connections in different levels and cross-attention based TD-MHCSA in the fast pathway. From Table~\ref{tab:Slowfast} we can obviously find that both lateral connections and TD-MHCSA improve the performance remarkably. This is because the former one brings more temporally fine-grained clues back to the slow pathway to alleviate rPPG information loss while the latter one leverages the cross -attention features to refine the redundant rPPG features in the fast pathway. The best configuration for PhysFormer++ is with high-level lateral connection and mid\&high-level TD-MHCSA.

%%\vspace{-0.2em}
\begin{figure}
\centering
\includegraphics[scale=0.18]{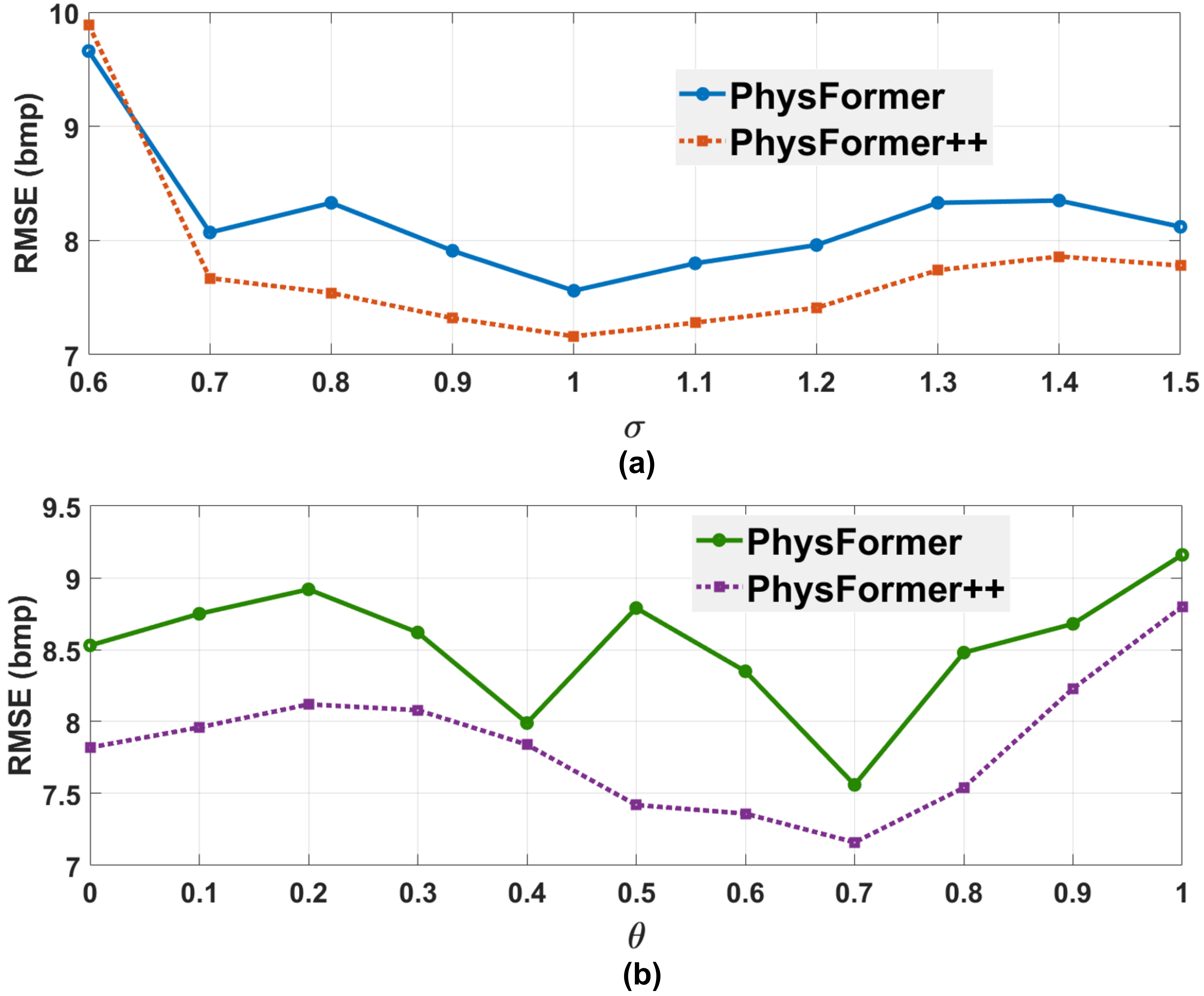}
%\vspace{-0.8em}
  \caption{\small{
 Impacts of the (a) $\sigma$ in label distribution learning for PhysFormer and PhysFormer++ and (b) $\theta$ in TD-MHSA, TD-MHCSA, and TD-MHPSA. }
  }
\label{fig:distribution}
%\vspace{-0.8em}
\end{figure}

\begin{figure}
\includegraphics[scale=0.35]{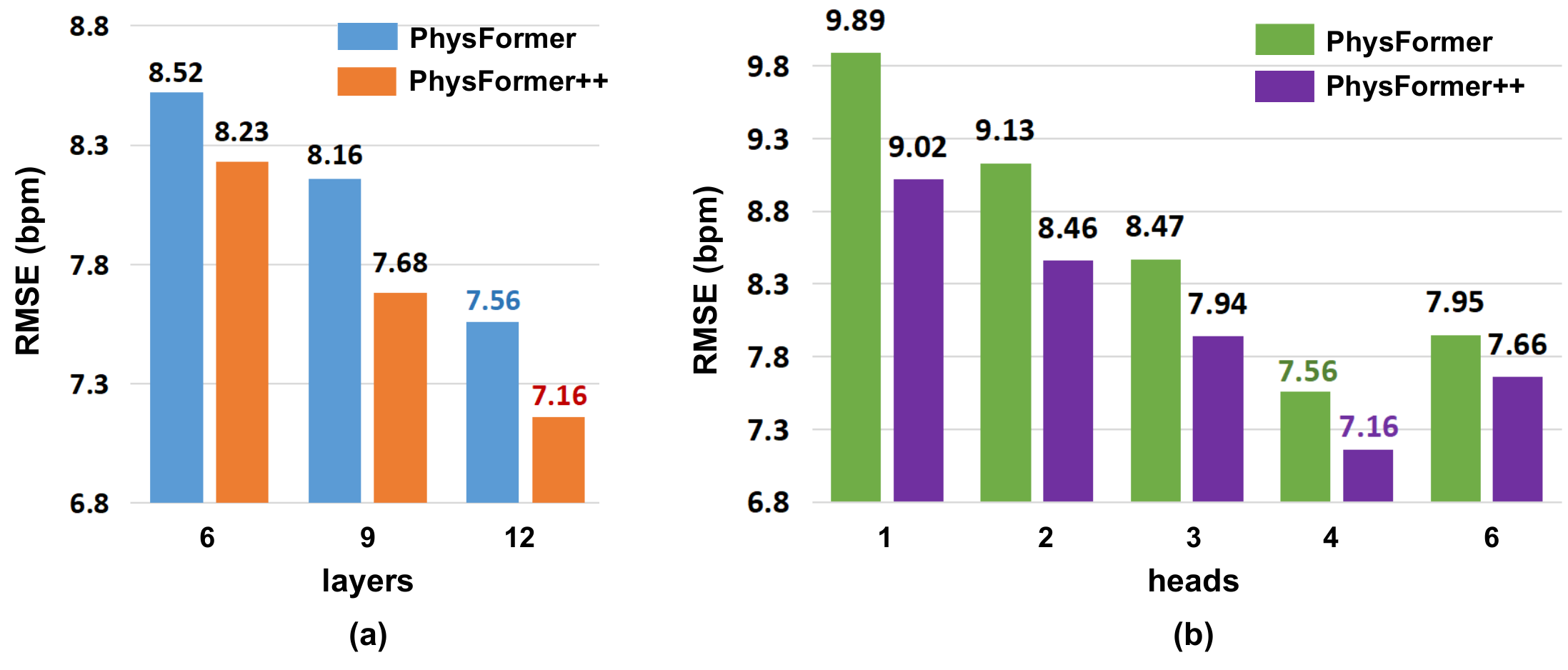}
%\vspace{-0.8em}
  \caption{\small{
 Ablation of the (a) layers and (b) heads in PhysFormer and PhysFormer++. }
  }
\label{fig:layerhead}
%\vspace{-0.8em}
\end{figure}

\vspace{0.4em}
\noindent\textbf{Impact of $\theta$ and layer/head numbers in the PhysFormer family}.\quad 
Hyperparameter $\theta$ tradeoffs the contribution of local temporal gradient information. As illustrated in Fig.~\ref{fig:distribution}(b), PhysFormer could achieve smaller RMSE when $\theta$=0.4 and 0.7 while PhysFormer++ obtains the best performance when $\theta$=0.7, indicating the importance of the normalized local temporal difference features for global spatio-temporal attention. We also investigate how the layer and head numbers influence the performance of PhysFormer and PhysFormer++. As shown in Fig.~\ref{fig:layerhead}(a), with deeper temporal transformer blocks, the RMSE are reduced progressively despite heavier computational cost. In terms of the impact of head numbers, it is clear to find from Fig.~\ref{fig:layerhead}(b) that the PhysFormer family with four heads performs the best while fewer heads lead to sharp performance drops.

\begin{table}
\centering

\caption{Ablation of dynamic loss in the frequency domain for PhysFormer and PhysFormer++. The temporal loss $\mathcal{L}_{time}$ is with fixed $\alpha$=0.1 here. `CE' and `LD' denote cross-entropy and label distribution, respectively.}

%\vspace{-0.9em}

%%\vspace{-0.8em}
\scalebox{0.78}{\begin{tabular}{l|c | c |c}
\toprule%\toprule[1pt]
Frequency loss & $\beta$ & Strategy & RMSE (bpm) $\downarrow$\\
\hline
\multicolumn{4}{c}{\textbf{PhysFormer}} \\
\hline
$\mathcal{L}_{\text{CE}}$ + $\mathcal{L}_{\text{LD}}$  & 1.0 & fixed & 8.48 \\
$\mathcal{L}_{\text{CE}}$ + $\mathcal{L}_{\text{LD}}$  & 5.0 & fixed & 8.86 \\
$\mathcal{L}_{\text{CE}}$ + $\mathcal{L}_{\text{LD}}$  & [1.0, 5.0] & linear & 8.37 \\
$\mathcal{L}_{\text{CE}}$ + $\mathcal{L}_{\text{LD}}$  & [1.0, 5.0] & exponential & \textbf{7.56} \\
$\mathcal{L}_{\text{CE}}$  & [1.0, 5.0] & exponential & 8.09 \\
$\mathcal{L}_{\text{LD}}$  & [1.0, 5.0] & exponential & 8.21 \\
$\mathcal{L}_{\text{LD}}$ (real distribution)  & [1.0, 5.0] & exponential & 8.72 \\
\hline
\multicolumn{4}{c}{\textbf{PhysFormer++}} \\
\hline
$\mathcal{L}_{\text{CE}}$ + $\mathcal{L}_{\text{LD}}$  & 1.0 & fixed & 7.98 \\
$\mathcal{L}_{\text{CE}}$ + $\mathcal{L}_{\text{LD}}$  & 5.0 & fixed & 8.54 \\
$\mathcal{L}_{\text{CE}}$ + $\mathcal{L}_{\text{LD}}$  & [1.0, 5.0] & linear & 8.13 \\
$\mathcal{L}_{\text{CE}}$ + $\mathcal{L}_{\text{LD}}$  & [1.0, 5.0] & exponential & \textbf{7.16} \\
$\mathcal{L}_{\text{CE}}$  & [1.0, 5.0] & exponential & 7.76 \\
$\mathcal{L}_{\text{LD}}$  & [1.0, 5.0] & exponential & 7.89 \\
$\mathcal{L}_{\text{LD}}$ (real distribution)  & [1.0, 5.0] & exponential & 8.67 \\
 
\bottomrule[0.8pt]

\end{tabular}}

\label{tab:ablation3}
%\vspace{-1.0em}
\end{table}

\vspace{0.4em}
\noindent\textbf{Impact of label distribution learning for the PhysFormer family}.\quad 
Besides the temporal loss $\mathcal{L}_{\text{time}}$ and frequency cross-entropy loss $\mathcal{L}_{\text{CE}}$, the ablations w/ and w/o label distribution loss $\mathcal{L}_{\text{LD}}$ are shown in the last four rows of Table~\ref{tab:ablation3}. Although the $\mathcal{L}_{\text{LD}}$ performs slightly worse (respective +0.12 and +0.13 bpm RMSE for PhysFormer and PhysFormer++) than $\mathcal{L}_{\text{CE}}$, the best performance can be achieved using both losses, indicating the effectiveness of explicit distribution constraints for extreme-frequency interference alleviation and adjacent label knowledgement propagation. It is interesting to find from the last two rows in both PhysFormer and PhysFormer++ that using real PSD distribution from ground truth PPG signals as $\mathbf{p}$, the performance is inferior due to the lack of an obvious peak in the distribution and partial noise. We can also find from the Fig.~\ref{fig:distribution}(a) that the $\sigma$ ranged from 0.9 to 1.2 for $\mathcal{L}_{\text{LD}}$ are suitable to achieve good performance.

%%\vspace{-0.2em}
\begin{figure}
\centering
\includegraphics[scale=0.18]{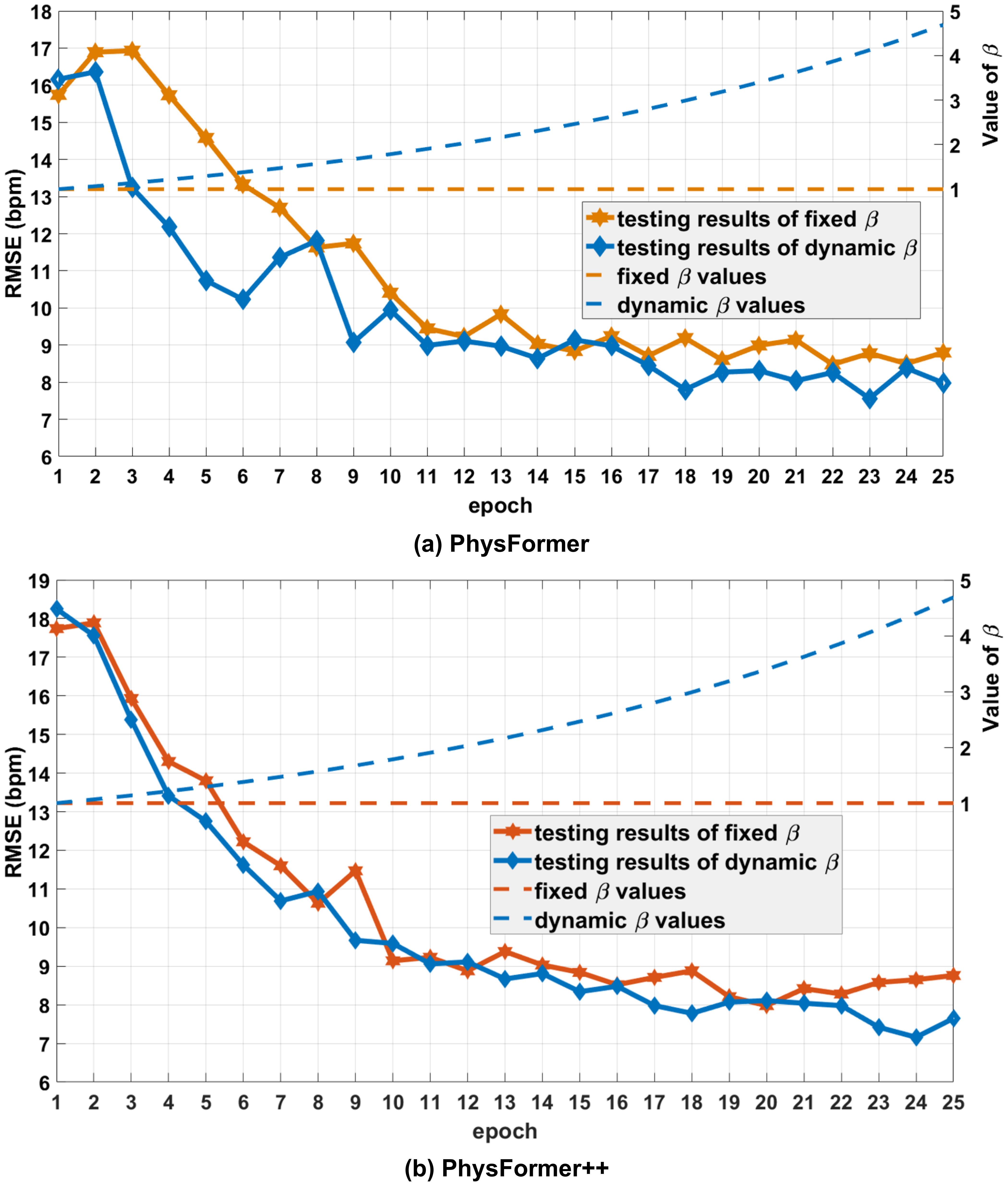}

%\vspace{-1.0em}
  \caption{\small{
 Testing results of fixed and dynamic frequency supervisions for (a) PhysFormer and (b) PhysFormer++ on the Fold-1 of VIPL-HR. }
  }
\label{fig:dynamicloss}
%\vspace{-1.2em}
\end{figure}

\vspace{0.4em}
\noindent\textbf{Impact of dynamic supervision for the PhysFormer family}.\quad 
Fig.~\ref{fig:dynamicloss} illustrates the testing performance of PhysFormer and PhysFormer++ on Fold-1 VIPL-HR when training with fixed and dynamic supervision. It is clear that with exponentially increased frequency loss, models in the blue curves converge faster and achieve smaller RMSE. We also compare several kinds of fixed and dynamic strategies in Table~\ref{tab:ablation3}. The results in the first four rows indicate 1) using fixed higher $\beta$ leads to poorer performance caused by the convergency difficulty; 2) models with the exponentially increased $\beta$ perform better than using linear increment.

\begin{figure*}
\centering
\includegraphics[scale=0.26]{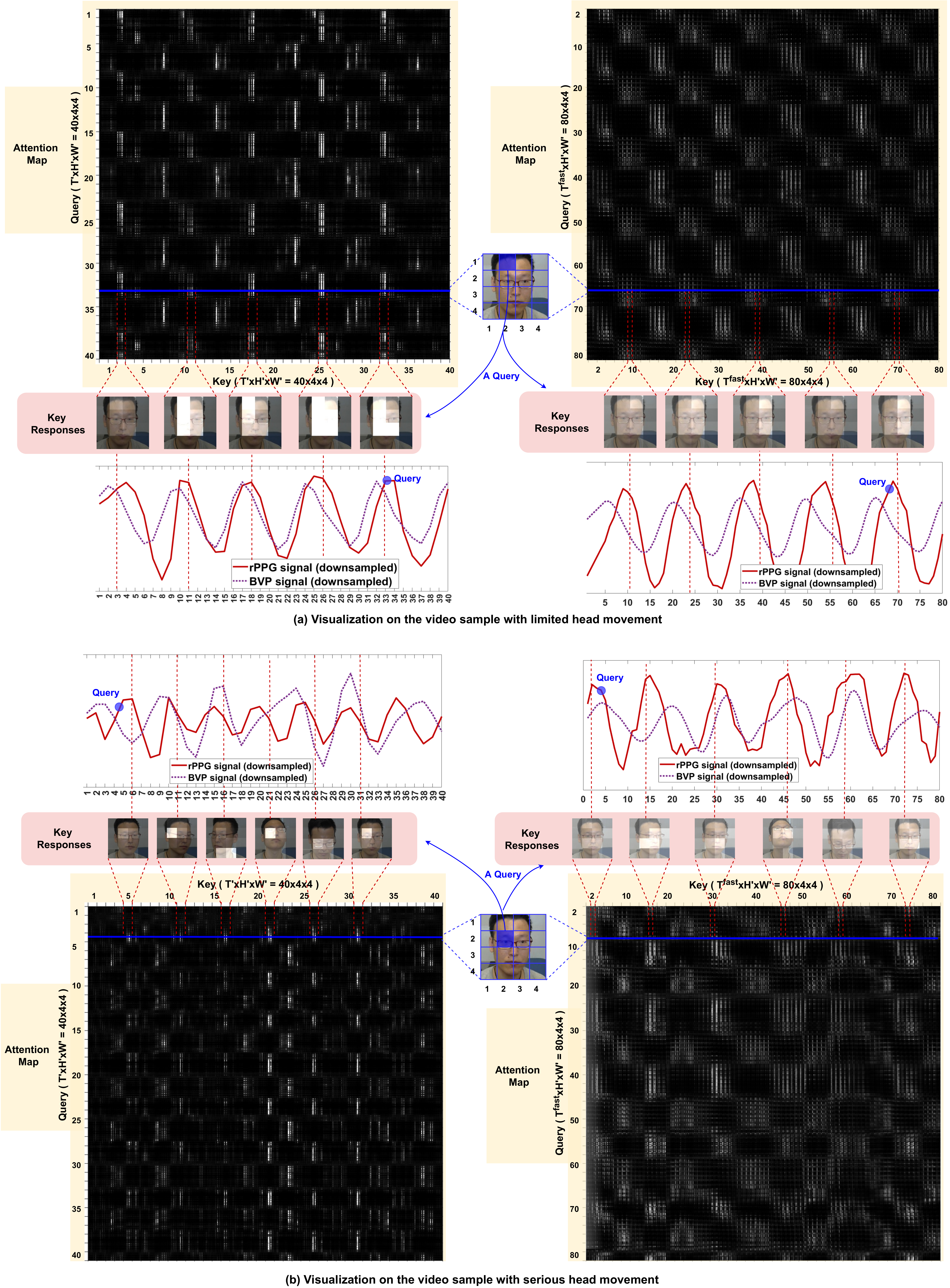}
%\vspace{-1.4em}
  \caption{\small{
 Visualization of the attention maps from (left) the 1st head in last TD-MHSA module of PhysFormer and (right) the 1st head in last TD-MHCSA module of the fast pathway in PhysFormer++. Given the 530th and 276th tube tokens in \textcolor{blue}{blue} as the query for the video samples with (a)  limited head movement and (b) serioud head movement, representative key responses are illustrated (the brighter, the more attentive). The predicted downsampled rPPG signals as well as the ground truth BVP signals are shown for temporal attention understanding. }
  }
\label{fig:visualization}
%\vspace{-0.8em}
\end{figure*}

\subsection{Efficiency Analysis}
 \label{sec:Efficiency}
Here we also investigate the computational cost\footnote{\href{https://pypi.org/project/thop/}{https://pypi.org/project/thop/}} compared with the baselines. The number of parameters and the multiply–accumulates (MACs) are shown in Table~\ref{tab:ResultsMMSE}. Despite huge  number of parameters, PhysFormer and PhysFormer++ are with smaller MACs compared with baselines PhysNet, TS-CAN, and AutoHR. Compared with PhysFormer, the PhysFormer++ introduces extra 2.76M paramters and 1.16G MACs. The inference time of one face clip 3x160x128x128 (CxTxHxW) for PhysFormer and PhysFormer++ on one V100 GPU is 29ms and 40ms, respectively. Despite slightly heavier, it can predict more accurate rPPG signals on both intra-dataset (-0.17 bpm RMSE on VIPL-HR) and cross-dataset (-0.21 bpm RMSE on MMSE-HR) testings. Towards efficient mobile-level rPPG applications, the computational cost of the proposed PhysFormer family is still unsatisfied. One potential future direction is to design more lightweight PhysFormer with advanced network quantization~\cite{lin2021fq} and binarization~\cite{qin2022bibert} techniques.

\begin{table}[t]\small
\centering
\caption{\small{Cross-dataset results with computational cost on MMSE-HR. The FLOPs are calculated with the video input size 3$\times$160$\times$128$\times$128 ($C\times T\times H\times W$) for PhysNet/ AutoHR/ PhysFormer/ PhysFormer++ while 3$\times$160$\times$96$\times$96 for TS-CAN/EfficientPhys.}} \label{tab:ResultsMMSE}

%\vspace{-0.8em}

\resizebox{0.47\textwidth}{!} {\begin{tabular}{l c c c c} 
 %\hline\hline
 \toprule
 Method & \begin{tabular}[c]{@{}c@{}} \#Param.\\(M)\end{tabular} & \begin{tabular}[c]{@{}c@{}} MACs\\(G)\end{tabular} & \begin{tabular}[c]{@{}c@{}}RMSE $\downarrow$ \\(bpm)\end{tabular}\\
 %\hline
 \midrule

PhysNet~\cite{yu2019remote1} & 0.73 & 65.19 & 13.25 \\
TS-CAN~\cite{liu2020multi} & 3.91 &  61.96 & 7.21  \\
AutoHR~\cite{yu2020autohr} & 0.99 & 189.22 & 5.87 \\
EfficientPhys-C~\cite{liu2021efficientphys} & 3.84 &  31.32 & 5.43 \\
\textbf{PhysFormer (Ours)} & 7.03 & 47.01 & \underline{5.36}\\
\textbf{PhysFormer++ (Ours)} & 9.79 & 49.85 & \textbf{5.15}\\

 \bottomrule[1pt]
 \end{tabular}}
\end{table}

\subsection{Visualization and Discussion}
 \label{sec:Analysis}

%\vspace{0.4em}
\noindent\textbf{Visualization of the self-attention map.}\quad 
We visualize the attention maps from the last TD-MHSA module of PhysFormer (left) and the last TD-MHCSA module in the fast pathway of PhysFormer++ (right) in Fig.~\ref{fig:visualization}. The x and y axes of the attention map indicate the attention confidence from key and query tube tokens, respectively. From the attention maps activated from the video sample with limited head movement in Fig.~\ref{fig:visualization}(a), we can easily find periodic or quasi-periodic responses along both axes, indicating the periodicity of the intrinsic rPPG features from PhysFormer and PhysFormer++. To be specific, given the 530th tube token (in blue) from the forehead (spatial face domain) and peak (temporal signal domain) locations as a query, the corresponding key responses are illustrated at the blue line in the attention map. On the one hand, it can be seen from the key responses that dominant spatial attentions focus on the facial skin regions and discard unrelated background. On the other hand, the temporal localizations of the key responses are around peak positions in the predicted rPPG signals. All these patterns are reasonable: 1) the forehead and cheek regions~\cite{verkruysse2008remote} have richer blood volume for rPPG measurement and are also reliable since these regions are less affected by facial muscle movements due to e.g., facial expressions, talking; and 2) rPPG signals from healthy people are usually periodic.

\begin{figure}[t]
\centering
\includegraphics[scale=0.48]{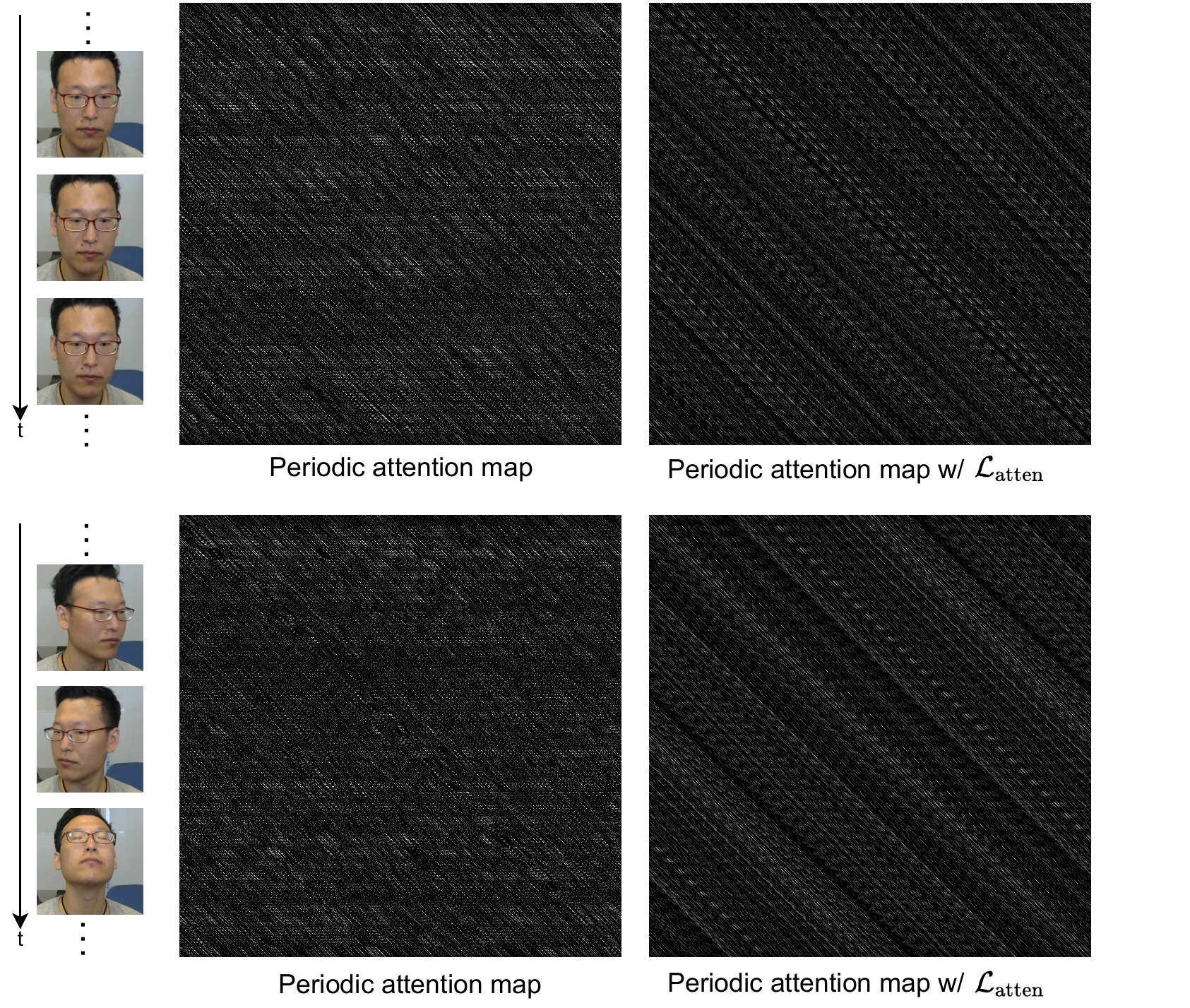}
%\vspace{-1.4em}
  \caption{\small{
 Visualization of the periodic attention maps from the 1st head in last TD-MHPSA module of the slow pathway in PhysFormer++. The top row show the periodic attention map from the facial video with limited head movement while the bottom one with serious head movement. }
  }
\label{fig:PeriodicMap}
%\vspace{-0.8em}
\end{figure}

We also visualize the attention maps from another video sample with serious head movement in Fig.~\ref{fig:visualization}(b). It can be observed from the left subfigure that the attentional response of PhysFormer is inaccurate (e.g., focusing on the neck region) when the head moves to the left. Another issue is that due to the large temporal token size ($T_{s}$=4) in the tokenization stage, the temporal rPPG clues might be partially discarded, resulting in the sensitivity about the head movement and the biased rPPG prediction (i.e., huge IBI gaps between the predicted rPPG and ground truth BVP signals). In contrast, it can be seen from the right subfigure in Fig.~\ref{fig:visualization}(b) that the attentional response and the predicted rPPG signal from PhysFormer++ are reliable, indicating the effectiveness of the SlowFast architecture and advanced attention modules. 

Overall, two limitations of the spatio-temporal attention could be concluded from Fig.~\ref{fig:visualization}. First, there are still some unexpected responses (e.g., continuous query tokens with similar key responses) in the attention map, which might introduce task-irrelevant noise and damage to the performance. Second, the temporal attentions are not accurate under serious head movement scenarios, and some are coarse with phase shifts.

\vspace{0.4em}
\noindent\textbf{Visualization of the periodic attention map.}\quad 
We also visualize the periodic attention map from the last TD-MHPSA module of PhysFormer++ in Fig.~\ref{fig:PeriodicMap}. It is interesting to find that the periodic attention maps from the PhysFormer++ 1) trained without $\mathcal{L}_{\text{atten}}$ are more arbitrary and easily influenced by the large head movement; and 2) trained with $\mathcal{L}_{\text{atten}}$ are more regular and keep the periodicity even under the scenarios with serious head movement. In other words, the proposed TD-MHPSA with attention loss $\mathcal{L}_{\text{atten}}$ enforces the PhysFormer++ to learn more periodic and robust attentional features from the face videos.

\begin{figure}[t]
\centering
\includegraphics[scale=0.42]{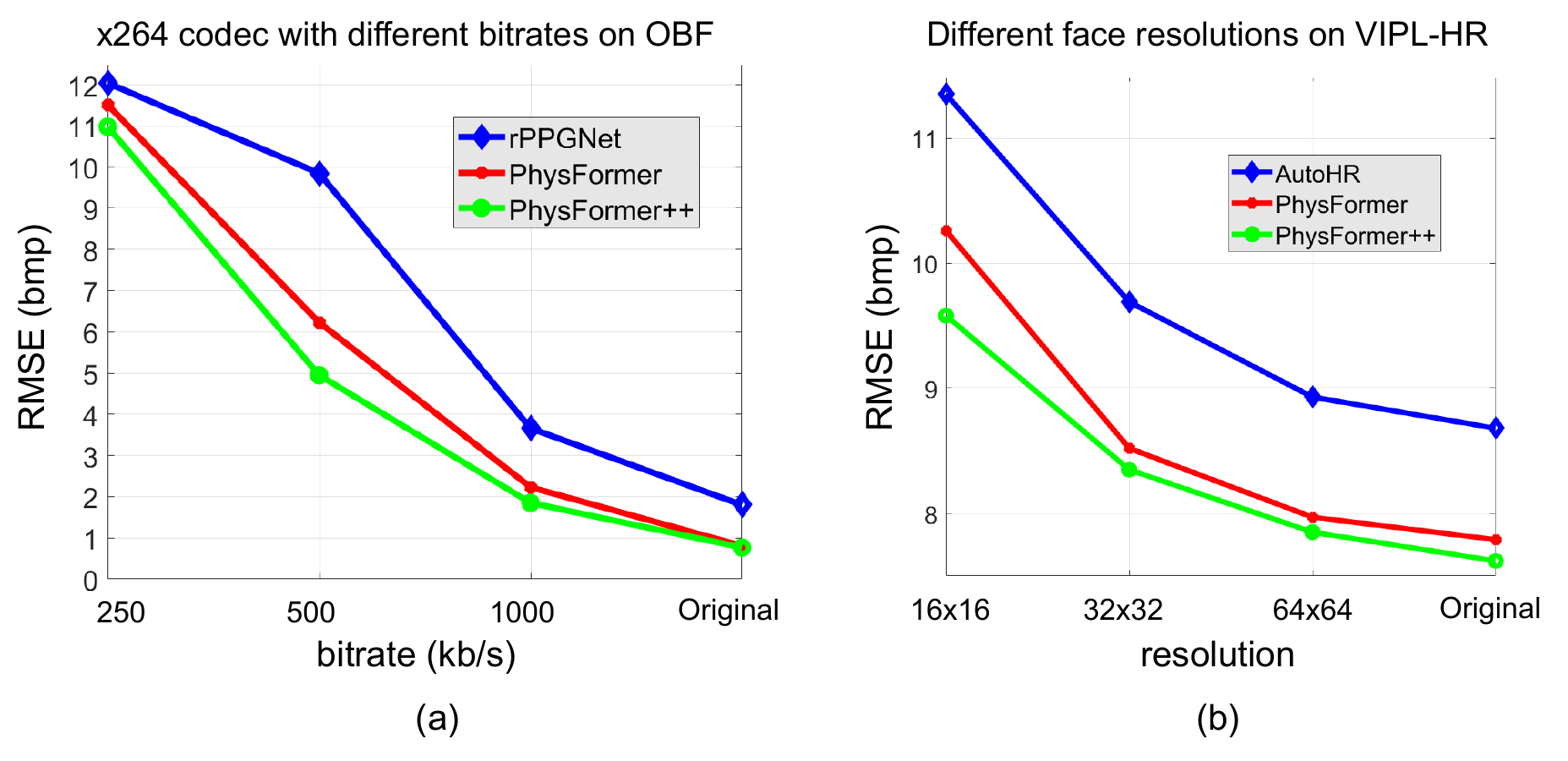}
%\vspace{-1.4em}
  \caption{\small{
HR results with different (a) compression bitrates on OBF, and (b) resolutions on VIPL-HR.} }
  
\label{fig:compression}
%\vspace{-0.8em}
\end{figure}

\vspace{0.4em}
\noindent\textbf{Evaluation under serious motion, video compression, and low resolution.}\quad 
In real-world scenarios, large head movement, high video compression rate and low face resolution usually introduce serious motion noises, compression artifacts and blurriness, respectively. All these corruptions and quality degradations make the rPPG measurement challenging. Here we evaluate the performance under these challenging scenarios. 

First, we evaluate the PhysFormer family under scenarios of large head movement (i.e., ‘v2’ and ‘v9’ samples) on VIPL-HR dataset. PhysFormer and PhysFormer++ achieve RMSE of 11.46 bpm and 10.25 bpm, respectively. In other words, with richer temporally contextual rPPG clues, the two-pathway SlowFast architecture in PhysFormer++ is more motion-robust. Note that there are still performance gaps between non-end-to-end method (e.g., RhythmNet~\cite{niu2019rhythmnet} with RMSE=9.4 bpm). 

Second, we evaluate the PhysFormer family on OBF with high compression rates (250/500/1000 kb/s) using x264 codec. The corresponding HR measurement results are illustrated in Fig.~\ref{fig:compression}(a). Compared with the rPPGNet~\cite{yu2019remote2}, the PhysFormer family performs significantly better when bitrates equal to 500 and 1000 kb/s. This might be because the spatio-temporal self-attention mechanism helps filter out the compression artifacts. However, all three methods perform poorly under extremely high compression situation (i.e., bitrate=250 kb/s).

Finally, we evaluate the PhysFormer family on VIPL-HR with different low-resolution settings to mimic the long-distance rPPG monitoring scenario. Specifically, bilinear interpolation is used to downsample the face frames to the sizes 16x16/32x32/64x64 first, and then upsample them back to 128x128. The HR measurement results are illustrated in Fig.~\ref{fig:compression}(b). Despite performance drops with lower face resolution for both AutoHR~\cite{yu2020autohr} and the PhysFormer family,  PhysFormer++ still achieves RMSE=9.58 bpm with the lowest (16x16) resolution setting.

\begin{table}[t]\small
\centering
\caption{\small{HR results (RMSE (bpm)) when training with different proportion of samples on VIPL-HR.}}\label{tab:fewerlabel}

%\vspace{-0.8em}

\resizebox{0.43\textwidth}{!} {\begin{tabular}{l c c c c} 
 %\hline\hline
 \toprule
 Method & 10\% & 50\% & 100\% \\
 %\hline
 \midrule

AutoHR~\cite{yu2020autohr} & 15.77 & \textbf{10.27} & 8.68 \\
\textbf{PhysFormer (Ours)} & 14.84 & 11.18 & 7.79\\
\textbf{PhysFormer++ (Ours)} & \textbf{13.92} & 10.29  & \textbf{7.62}\\

 \bottomrule[1pt]
 \end{tabular}}
\end{table}

\vspace{0.4em}
\noindent\textbf{Training with fewer samples.}\quad 
Since end-to-end deep models (e.g., CNNs and transformers) are data hungry, here we investigate three methods (AutoHR~\cite{yu2020autohr}, PhysFormer and PhysFormer++) under conditions of fewer training samples. As shown in Table~\ref{tab:fewerlabel}, when training with only 10\% or 50\% samples, all these three methods obtain poor RMSE performance ($\textgreater$10 bpm). Another observation is that, compared with pure CNN-based AutoHR, the proposed PhysFormer++ still achieves on par or better performance with fewer training samples. It indicates that the proposed transformer architectures can learn CNN-comparable rPPG representation even with limited data.

%\vspace{-0.5em}
\section{Conclusions}
%\vspace{-0.1em}
\label{sec:conclusion}

In this paper, we propose two end-to-end video transformer architectures, namely PhysFormer and PhysFormer++, for remote physiological measurement. With temporal difference transformer and elaborate supervisions, the PhysFormer family is able to achieve superior performance on benchmark datasets on both intra- and cross-testings. Comprehensive ablation studies as well as visualization analysis demonstrate the effectiveness of the proposed methods. In the future, it is potential to explore more accurate yet efficient spatio-temporal self-attention mechanism especially for long sequence rPPG monitoring. Besides the rPPG measurement task, we will investigate the effectiveness of the proposed temporal different transformer for broader fine-grained or periodic video understanding tasks in computer vision (e.g., video action recognition and repetition counting).

% 7376257 params   7.03 M
% 50481262240 FLOPs  47.01 G
% inputs_temp = torch.randn(1, 3, 160, 128, 128)
%  flops, params = profile(model_2DViT, 
% inputs=(inputs_temp, gra_sharp))

\vspace{0.4em}
\noindent\textbf{Acknowledgment}  \quad This work was supported by the Academy of Finland (Academy Professor project
EmotionAI with grant numbers 336116 and 345122, and ICT2023 project with grant number 345948), the National Natural Science Foundation of China (Grant No. 62002283), the EPSRC grant: Turing AI Fellowship: EP/W002981/1, EPSRC/MURI grant EP/N019474/1. We would also like to thank the Royal Academy of Engineering and FiveAI. As well, the authors wish to acknowledge CSC-IT Center for Science, Finland, for computational resources.

\bibliographystyle{apalike}
\bibliography{sn-bibliography}% common bib file

\begin{thebibliography}{}

\bibitem[Arnab et~al., 2021]{arnab2021vivit}
Arnab, A., Dehghani, M., Heigold, G., Sun, C., Lu{\v{c}}i{\'c}, M., and Schmid,
  C. (2021).
\newblock Vivit: A video vision transformer.
\newblock {\em ICCV}.

\bibitem[Bengio et~al., 2009]{bengio2009curriculum}
Bengio, Y., Louradour, J., Collobert, R., and Weston, J. (2009).
\newblock Curriculum learning.
\newblock In {\em ICML}.

\bibitem[Bertasius et~al., 2021]{bertasius2021space}
Bertasius, G., Wang, H., and Torresani, L. (2021).
\newblock Is space-time attention all you need for video understanding?
\newblock {\em arXiv:2102.05095}.

\bibitem[Bulat et~al., 2021]{bulat2021space}
Bulat, A., Perez-Rua, J.-M., Sudhakaran, S., Martinez, B., and Tzimiropoulos,
  G. (2021).
\newblock Space-time mixing attention for video transformer.
\newblock {\em NeurIPS}.

\bibitem[Cao et~al., 2021]{cao2021video}
Cao, J., Li, Y., Zhang, K., and Van~Gool, L. (2021).
\newblock Video super-resolution transformer.
\newblock {\em arXiv:2106.06847}.

\bibitem[Carion et~al., 2020]{carion2020end}
Carion, N., Massa, F., Synnaeve, G., Usunier, N., Kirillov, A., and Zagoruyko,
  S. (2020).
\newblock End-to-end object detection with transformers.
\newblock In {\em ECCV}.

\bibitem[Carreira and Zisserman, 2017]{carreira2017quo}
Carreira, J. and Zisserman, A. (2017).
\newblock Quo vadis, action recognition? a new model and the kinetics dataset.
\newblock In {\em CVPR}.

\bibitem[Chen et~al., 2021a]{chen2021crossvit}
Chen, C.-F., Fan, Q., and Panda, R. (2021a).
\newblock Crossvit: Cross-attention multi-scale vision transformer for image
  classification.
\newblock {\em ICCV}.

\bibitem[Chen et~al., 2021b]{chen2021aniformer}
Chen, H., Tang, H., Sebe, N., and Zhao, G. (2021b).
\newblock Aniformer: Data-driven 3d animation with transformer.
\newblock {\em BMVC}.

\bibitem[Chen et~al., 2022]{chen2021geometry}
Chen, H., Tang, H., Yu, Z., Sebe, N., and Zhao, G. (2022).
\newblock Geometry-contrastive transformer for generalized 3d pose transfer.
\newblock In {\em AAAI}.

\bibitem[Chen and McDuff, 2018]{chen2018deepphys}
Chen, W. and McDuff, D. (2018).
\newblock Deepphys: Video-based physiological measurement using convolutional
  attention networks.
\newblock In {\em ECCV}.

\bibitem[Chen et~al., 2018]{chen2018video}
Chen, X., Cheng, J., Song, R., Liu, Y., Ward, R., and Wang, Z.~J. (2018).
\newblock Video-based heart rate measurement: Recent advances and future
  prospects.
\newblock {\em IEEE Transactions on Instrumentation and Measurement}.

\bibitem[Cho et~al., 2014]{cho2014properties}
Cho, K., Van~Merri{\"e}nboer, B., Bahdanau, D., and Bengio, Y. (2014).
\newblock On the properties of neural machine translation: Encoder-decoder
  approaches.
\newblock {\em arXiv preprint arXiv:1409.1259}.

\bibitem[De~Haan and Jeanne, 2013]{de2013robust}
De~Haan, G. and Jeanne, V. (2013).
\newblock Robust pulse rate from chrominance-based rppg.
\newblock {\em IEEE Transactions on Biomedical Engineering}.

\bibitem[Ding et~al., 2021]{ding2021hr}
Ding, M., Lian, X., Yang, L., Wang, P., Jin, X., Lu, Z., and Luo, P. (2021).
\newblock Hr-nas: Searching efficient high-resolution neural architectures with
  lightweight transformers.
\newblock In {\em CVPR}.

\bibitem[Dosovitskiy et~al., 2021]{dosovitskiy2020image}
Dosovitskiy, A., Beyer, L., Kolesnikov, A., Weissenborn, D., Zhai, X.,
  Unterthiner, T., Dehghani, M., Minderer, M., Heigold, G., Gelly, S., et~al.
  (2021).
\newblock An image is worth 16x16 words: Transformers for image recognition at
  scale.
\newblock {\em ICLR}.

\bibitem[Fan et~al., 2021]{fan2021multiscale}
Fan, H., Xiong, B., Mangalam, K., Li, Y., Yan, Z., Malik, J., and
  Feichtenhofer, C. (2021).
\newblock Multiscale vision transformers.
\newblock {\em ICCV}.

\bibitem[Feichtenhofer et~al., 2019]{feichtenhofer2019slowfast}
Feichtenhofer, C., Fan, H., Malik, J., and He, K. (2019).
\newblock Slowfast networks for video recognition.
\newblock In {\em CVPR}.

\bibitem[Gao et~al., 2017]{gao2017deep}
Gao, B.-B., Xing, C., Xie, C.-W., Wu, J., and Geng, X. (2017).
\newblock Deep label distribution learning with label ambiguity.
\newblock {\em IEEE TIP}.

\bibitem[Gao et~al., 2018]{gao2018age}
Gao, B.-B., Zhou, H.-Y., Wu, J., and Geng, X. (2018).
\newblock Age estimation using expectation of label distribution learning.
\newblock In {\em IJCAI}.

\bibitem[Geng et~al., 2013]{geng2013facial}
Geng, X., Yin, C., and Zhou, Z.-H. (2013).
\newblock Facial age estimation by learning from label distributions.
\newblock {\em IEEE TPAMI}.

\bibitem[Gideon and Stent, 2021]{gideon2021way}
Gideon, J. and Stent, S. (2021).
\newblock The way to my heart is through contrastive learning: Remote
  photoplethysmography from unlabelled video.
\newblock In {\em ICCV}.

\bibitem[Girdhar et~al., 2019]{girdhar2019video}
Girdhar, R., Carreira, J., Doersch, C., and Zisserman, A. (2019).
\newblock Video action transformer network.
\newblock In {\em CVPR}.

\bibitem[Han et~al., 2020]{han2020survey}
Han, K., Wang, Y., Chen, H., Chen, X., Guo, J., Liu, Z., Tang, Y., Xiao, A.,
  Xu, C., Xu, Y., et~al. (2020).
\newblock A survey on visual transformer.
\newblock {\em arXiv:2012.12556}.

\bibitem[Han et~al., 2021]{han2021transformer}
Han, K., Xiao, A., Wu, E., Guo, J., Xu, C., and Wang, Y. (2021).
\newblock Transformer in transformer.
\newblock {\em arXiv:2103.00112}.

\bibitem[Hassani et~al., 2021]{hassani2021escaping}
Hassani, A., Walton, S., Shah, N., Abuduweili, A., Li, J., and Shi, H. (2021).
\newblock Escaping the big data paradigm with compact transformers.
\newblock In {\em CVPRW}.

\bibitem[He et~al., 2016]{he2016deep}
He, K., Zhang, X., Ren, S., and Sun, J. (2016).
\newblock Deep residual learning for image recognition.
\newblock In {\em CVPR}.

\bibitem[He et~al., 2021]{he2021transreid}
He, S., Luo, H., Wang, P., Wang, F., Li, H., and Jiang, W. (2021).
\newblock Transreid: Transformer-based object re-identification.
\newblock {\em ICCV}.

\bibitem[Hsu et~al., 2017]{hsu2017deep}
Hsu, G.-S., Ambikapathi, A., and Chen, M.-S. (2017).
\newblock Deep learning with time-frequency representation for pulse estimation
  from facial videos.
\newblock In {\em IJCB}.

\bibitem[Huang et~al., 2019]{huang2018music}
Huang, C.-Z.~A., Vaswani, A., Uszkoreit, J., Shazeer, N., Simon, I., Hawthorne,
  C., Dai, A.~M., Hoffman, M.~D., Dinculescu, M., and Eck, D. (2019).
\newblock Music transformer.
\newblock In {\em ICLR}.

\bibitem[Kazakos et~al., 2021]{kazakos2021slow}
Kazakos, E., Nagrani, A., Zisserman, A., and Damen, D. (2021).
\newblock Slow-fast auditory streams for audio recognition.
\newblock In {\em ICASSP}.

\bibitem[Khan et~al., 2021]{khan2021transformers}
Khan, S., Naseer, M., Hayat, M., Zamir, S.~W., Khan, F.~S., and Shah, M.
  (2021).
\newblock Transformers in vision: A survey.
\newblock {\em arXiv:2101.01169}.

\bibitem[Lam and Kuno, 2015]{lam2015robust}
Lam, A. and Kuno, Y. (2015).
\newblock Robust heart rate measurement from video using select random patches.
\newblock In {\em ICCV}.

\bibitem[Lee et~al., 2020]{lee2020meta}
Lee, E., Chen, E., and Lee, C.-Y. (2020).
\newblock Meta-rppg: Remote heart rate estimation using a transductive
  meta-learner.
\newblock In {\em ECCV}.

\bibitem[Li et~al., 2018]{li2018obf}
Li, X., Alikhani, I., Shi, J., Seppanen, T., Junttila, J., Majamaa-Voltti, K.,
  Tulppo, M., and Zhao, G. (2018).
\newblock The obf database: A large face video database for remote
  physiological signal measurement and atrial fibrillation detection.
\newblock In {\em FG}.

\bibitem[Li et~al., 2014]{li2014remote}
Li, X., Chen, J., Zhao, G., and Pietikainen, M. (2014).
\newblock Remote heart rate measurement from face videos under realistic
  situations.
\newblock In {\em CVPR}.

\bibitem[Lin et~al., 2019]{lin2019tsm}
Lin, J., Gan, C., and Han, S. (2019).
\newblock Tsm: Temporal shift module for efficient video understanding.
\newblock In {\em CVPR}.

\bibitem[Lin et~al., 2021a]{lin2021survey}
Lin, T., Wang, Y., Liu, X., and Qiu, X. (2021a).
\newblock A survey of transformers.
\newblock {\em arXiv:2106.04554}.

\bibitem[Lin et~al., 2021b]{lin2021fq}
Lin, Y., Zhang, T., Sun, P., Li, Z., and Zhou, S. (2021b).
\newblock Fq-vit: Fully quantized vision transformer without retraining.
\newblock {\em arXiv preprint arXiv:2111.13824}.

\bibitem[Liu et~al., 2021a]{liu2021fuseformer}
Liu, R., Deng, H., Huang, Y., Shi, X., Lu, L., Sun, W., Wang, X., Dai, J., and
  Li, H. (2021a).
\newblock Fuseformer: Fusing fine-grained information in transformers for video
  inpainting.
\newblock In {\em ICCV}.

\bibitem[Liu et~al., 2020]{liu2020multi}
Liu, X., Fromm, J., Patel, S., and McDuff, D. (2020).
\newblock Multi-task temporal shift attention networks for on-device
  contactless vitals measurement.
\newblock {\em NeurIPS}.

\bibitem[Liu et~al., 2021b]{liu2021efficientphys}
Liu, X., Hill, B.~L., Jiang, Z., Patel, S., and McDuff, D. (2021b).
\newblock Efficientphys: Enabling simple, fast and accurate camera-based vitals
  measurement.
\newblock {\em arXiv:2110.04447}.

\bibitem[Liu et~al., 2021c]{liu2021camera}
Liu, X., Patel, S., and McDuff, D. (2021c).
\newblock Camera-based physiological sensing: Challenges and future directions.
\newblock {\em arXiv:2110.13362}.

\bibitem[Liu et~al., 2021d]{liu2021end}
Liu, X., Wang, Q., Hu, Y., Tang, X., Bai, S., and Bai, X. (2021d).
\newblock End-to-end temporal action detection with transformer.
\newblock {\em arXiv:2106.10271}.

\bibitem[Liu et~al., 2021e]{liu2021swin}
Liu, Z., Lin, Y., Cao, Y., Hu, H., Wei, Y., Zhang, Z., Lin, S., and Guo, B.
  (2021e).
\newblock Swin transformer: Hierarchical vision transformer using shifted
  windows.
\newblock {\em ICCV}.

\bibitem[Liu et~al., 2021f]{liu2021video}
Liu, Z., Ning, J., Cao, Y., Wei, Y., Zhang, Z., Lin, S., and Hu, H. (2021f).
\newblock Video swin transformer.
\newblock {\em arXiv:2106.13230}.

\bibitem[Lu and Han, 2021]{lu2021hr}
Lu, H. and Han, H. (2021).
\newblock Nas-hr: Neural architecture search for heart rate estimation from
  face videos.
\newblock {\em Virtual Reality \& Intelligent Hardware}.

\bibitem[Lu et~al., 2021]{lu2021dual}
Lu, H., Han, H., and Zhou, S.~K. (2021).
\newblock Dual-gan: Joint bvp and noise modeling for remote physiological
  measurement.
\newblock In {\em CVPR}.

\bibitem[Magdalena~Nowara et~al., 2018]{magdalena2018sparseppg}
Magdalena~Nowara, E., Marks, T.~K., Mansour, H., and Veeraraghavan, A. (2018).
\newblock Sparseppg: Towards driver monitoring using camera-based vital signs
  estimation in near-infrared.
\newblock In {\em CVPRW}.

\bibitem[Neimark et~al., 2021]{neimark2021video}
Neimark, D., Bar, O., Zohar, M., and Asselmann, D. (2021).
\newblock Video transformer network.
\newblock {\em arXiv:2102.00719}.

\bibitem[Niu et~al., 2017]{niu2017continuous}
Niu, X., Han, H., Shan, S., and Chen, X. (2017).
\newblock Continuous heart rate measurement from face: A robust rppg approach
  with distribution learning.
\newblock In {\em IJCB}.

\bibitem[Niu et~al., 2018]{niu2018synrhythm}
Niu, X., Han, H., Shan, S., and Chen, X. (2018).
\newblock Synrhythm: Learning a deep heart rate estimator from general to
  specific.
\newblock In {\em ICPR}.

\bibitem[Niu et~al., 2019a]{niu2019rhythmnet}
Niu, X., Shan, S., Han, H., and Chen, X. (2019a).
\newblock Rhythmnet: End-to-end heart rate estimation from face via
  spatial-temporal representation.
\newblock {\em IEEE TIP}.

\bibitem[Niu et~al., 2020]{niu2020video}
Niu, X., Yu, Z., Han, H., Li, X., Shan, S., and Zhao, G. (2020).
\newblock Video-based remote physiological measurement via cross-verified
  feature disentangling.
\newblock In {\em ECCV}, pages 295--310. Springer.

\bibitem[Niu et~al., 2019b]{niu2019robust}
Niu, X., Zhao, X., Han, H., Das, A., Dantcheva, A., Shan, S., and Chen, X.
  (2019b).
\newblock Robust remote heart rate estimation from face utilizing
  spatial-temporal attention.
\newblock In {\em FG}.

\bibitem[Nowara et~al., 2021]{nowara2021benefit}
Nowara, E.~M., McDuff, D., and Veeraraghavan, A. (2021).
\newblock The benefit of distraction: Denoising camera-based physiological
  measurements using inverse attention.
\newblock In {\em ICCV}.

\bibitem[Poh et~al., 2010a]{poh2010advancements}
Poh, M.-Z., McDuff, D., and Picard, R. (2010a).
\newblock Advancements in noncontact, multiparameter physiological measurements
  using a webcam.
\newblock {\em IEEE transactions on biomedical engineering}.

\bibitem[Poh et~al., 2010b]{poh2010non}
Poh, M.-Z., McDuff, D.~J., and Picard, R.~W. (2010b).
\newblock Non-contact, automated cardiac pulse measurements using video imaging
  and blind source separation.
\newblock {\em Optics express}.

\bibitem[Qin et~al., 2022]{qin2022bibert}
Qin, H., Ding, Y., Zhang, M., Yan, Q., Liu, A., Dang, Q., Liu, Z., and Liu, X.
  (2022).
\newblock Bibert: Accurate fully binarized bert.
\newblock In {\em ICLR}.

\bibitem[Qiu et~al., 2018]{qiu2018evm}
Qiu, Y., Liu, Y., Arteaga-Falconi, J., Dong, H., and El~Saddik, A. (2018).
\newblock Evm-cnn: Real-time contactless heart rate estimation from facial
  video.
\newblock {\em IEEE TMM}.

\bibitem[Revanur et~al., 2022]{revanur2022instantaneous}
Revanur, A., Dasari, A., Tucker, C.~S., and Jeni, L.~A. (2022).
\newblock Instantaneous physiological estimation using video transformers.
\newblock {\em arXiv preprint arXiv:2202.12368}.

\bibitem[Shaw et~al., 2018]{shaw2018self}
Shaw, P., Uszkoreit, J., and Vaswani, A. (2018).
\newblock Self-attention with relative position representations.
\newblock {\em arXiv preprint arXiv:1803.02155}.

\bibitem[Soleymani et~al., 2011]{soleymani2011multimodal}
Soleymani, M., Lichtenauer, J., Pun, T., and Pantic, M. (2011).
\newblock A multimodal database for affect recognition and implicit tagging.
\newblock {\em IEEE transactions on affective computing}.

\bibitem[{\v{S}}petl{\'\i}k et~al., 2018]{vspetlik2018visual}
{\v{S}}petl{\'\i}k, R., Franc, V., and Matas, J. (2018).
\newblock Visual heart rate estimation with convolutional neural network.
\newblock In {\em BMVC}.

\bibitem[Touvron et~al., 2021]{touvron2021training}
Touvron, H., Cord, M., Douze, M., Massa, F., Sablayrolles, A., and J{\'e}gou,
  H. (2021).
\newblock Training data-efficient image transformers \& distillation through
  attention.
\newblock In {\em ICML}.

\bibitem[Tulyakov et~al., 2016]{tulyakov2016self}
Tulyakov, S., Alameda-Pineda, X., Ricci, E., Yin, L., Cohn, J.~F., and Sebe, N.
  (2016).
\newblock Self-adaptive matrix completion for heart rate estimation from face
  videos under realistic conditions.
\newblock In {\em CVPR}.

\bibitem[Vaswani et~al., 2017]{vaswani2017attention}
Vaswani, A., Shazeer, N., Parmar, N., Uszkoreit, J., Jones, L., Gomez, A.~N.,
  Kaiser, {\L}., and Polosukhin, I. (2017).
\newblock Attention is all you need.
\newblock In {\em NIPS}.

\bibitem[Verkruysse et~al., 2008]{verkruysse2008remote}
Verkruysse, W., Svaasand, L.~O., and Nelson, J.~S. (2008).
\newblock Remote plethysmographic imaging using ambient light.
\newblock {\em Optics express}.

\bibitem[Wang et~al., 2021a]{wang2021temporal}
Wang, L., Yang, H., Wu, W., Yao, H., and Huang, H. (2021a).
\newblock Temporal action proposal generation with transformers.
\newblock {\em arXiv:2105.12043}.

\bibitem[Wang et~al., 2017]{wang2017algorithmic}
Wang, W., den Brinker, A.~C., Stuijk, S., and de~Haan, G. (2017).
\newblock Algorithmic principles of remote ppg.
\newblock {\em IEEE Transactions on Biomedical Engineering}.

\bibitem[Wang et~al., 2021b]{wang2021pyramid}
Wang, W., Xie, E., Li, X., Fan, D.-P., Song, K., Liang, D., Lu, T., Luo, P.,
  and Shao, L. (2021b).
\newblock Pyramid vision transformer: A versatile backbone for dense prediction
  without convolutions.
\newblock {\em ICCV}.

\bibitem[Wu et~al., 2021]{wu2021rethinking}
Wu, K., Peng, H., Chen, M., Fu, J., and Chao, H. (2021).
\newblock Rethinking and improving relative position encoding for vision
  transformer.
\newblock In {\em ICCV}.

\bibitem[Xiao et~al., 2021]{xiao2021early}
Xiao, T., Singh, M., Mintun, E., Darrell, T., Doll{\'a}r, P., and Girshick, R.
  (2021).
\newblock Early convolutions help transformers see better.
\newblock {\em NeurIPS}.

\bibitem[Xu et~al., 2021]{xu2021long}
Xu, M., Xiong, Y., Chen, H., Li, X., Xia, W., Tu, Z., and Soatto, S. (2021).
\newblock Long short-term transformer for online action detection.
\newblock {\em arXiv:2107.03377}.

\bibitem[Yu et~al., 2020]{yu2020autohr}
Yu, Z., Li, X., Niu, X., Shi, J., and Zhao, G. (2020).
\newblock Autohr: A strong end-to-end baseline for remote heart rate
  measurement with neural searching.
\newblock {\em IEEE SPL}.

\bibitem[Yu et~al., 2021a]{yu2021transrppg}
Yu, Z., Li, X., Wang, P., and Zhao, G. (2021a).
\newblock Transrppg: Remote photoplethysmography transformer for 3d mask face
  presentation attack detection.
\newblock {\em IEEE SPL}.

\bibitem[Yu et~al., 2019a]{yu2019remote1}
Yu, Z., Li, X., and Zhao, G. (2019a).
\newblock Remote photoplethysmograph signal measurement from facial videos
  using spatio-temporal networks.
\newblock In {\em BMVC}.

\bibitem[Yu et~al., 2021b]{yu2021facial}
Yu, Z., Li, X., and Zhao, G. (2021b).
\newblock Facial-video-based physiological signal measurement: Recent advances
  and affective applications.
\newblock {\em IEEE Signal Processing Magazine}.

\bibitem[Yu et~al., 2019b]{yu2019remote2}
Yu, Z., Peng, W., Li, X., Hong, X., and Zhao, G. (2019b).
\newblock Remote heart rate measurement from highly compressed facial videos:
  an end-to-end deep learning solution with video enhancement.
\newblock In {\em ICCV}.

\bibitem[Yu et~al., 2021c]{yu2021deep}
Yu, Z., Qin, Y., Li, X., Zhao, C., Lei, Z., and Zhao, G. (2021c).
\newblock Deep learning for face anti-spoofing: a survey.
\newblock {\em arXiv:2106.14948}.

\bibitem[Yu et~al., 2022]{yu2021physformer}
Yu, Z., Shen, Y., Shi, J., Zhao, H., Torr, P., and Zhao, G. (2022).
\newblock Physformer: Facial video-based physiological measurement with
  temporal difference transformer.
\newblock In {\em CVPR}.

\bibitem[Yu et~al., 2021d]{yu2021searching}
Yu, Z., Zhou, B., Wan, J., Wang, P., Chen, H., Liu, X., Li, S.~Z., and Zhao, G.
  (2021d).
\newblock Searching multi-rate and multi-modal temporal enhanced networks for
  gesture recognition.
\newblock {\em IEEE TIP}.

\bibitem[Yuan et~al., 2021]{yuan2021tokens}
Yuan, L., Chen, Y., Wang, T., Yu, W., Shi, Y., Jiang, Z., Tay, F.~E., Feng, J.,
  and Yan, S. (2021).
\newblock Tokens-to-token vit: Training vision transformers from scratch on
  imagenet.
\newblock {\em arXiv:2101.11986}.

\bibitem[Zeng et~al., 2020]{zeng2020learning}
Zeng, Y., Fu, J., and Chao, H. (2020).
\newblock Learning joint spatial-temporal transformations for video inpainting.
\newblock In {\em ECCV}.

\bibitem[Zhang et~al., 2016]{zhang2016joint}
Zhang, K., Zhang, Z., Li, Z., and Qiao, Y. (2016).
\newblock Joint face detection and alignment using multitask cascaded
  convolutional networks.
\newblock {\em IEEE SPL}.

\bibitem[Zhao et~al., 2021]{zhao2021tuber}
Zhao, J., Li, X., Liu, C., Bing, S., Chen, H., Snoek, C.~G., and Tighe, J.
  (2021).
\newblock Tuber: Tube-transformer for action detection.
\newblock {\em arXiv:2104.00969}.

\bibitem[Zheng et~al., 2021]{zheng2021rethinking}
Zheng, S., Lu, J., Zhao, H., Zhu, X., Luo, Z., Wang, Y., Fu, Y., Feng, J.,
  Xiang, T., Torr, P.~H., et~al. (2021).
\newblock Rethinking semantic segmentation from a sequence-to-sequence
  perspective with transformers.
\newblock In {\em CVPR}.

\end{thebibliography}



\end{document}